%% file: main.tex
\documentclass[journal]{IEEEtran}
\usepackage{amsmath,amsfonts}
\usepackage{algorithmic}
\usepackage{algorithm}
\usepackage{array}
\usepackage[caption=false,font=normalsize,labelfont=sf,textfont=sf]{subfig}
\usepackage{textcomp}
\usepackage{stfloats}
\usepackage{url}
\usepackage{verbatim}
\usepackage{graphicx}
\usepackage[style=numeric]{biblatex}
\newcommand{\TokIn}{$T_{input}$}
\newcommand{\TokOut}{$T_{out}$}

\begin{document}

\title{From Prompts to Power: Measuring the Energy Footprint of LLM Inference}

\author{
    \IEEEauthorblockN{Francisco Caravaca\IEEEauthorrefmark{1}, Ángel Cuevas\IEEEauthorrefmark{1}\IEEEauthorrefmark{2}, Rubén Cuevas\IEEEauthorrefmark{1} \IEEEauthorrefmark{2} \IEEEauthorrefmark{3}}
    
    \IEEEauthorblockA{\IEEEauthorrefmark{1}Universidad Carlos III de Madrid}
    \IEEEauthorblockA{\IEEEauthorrefmark{2}Hiili S.L.}
    \IEEEauthorblockA{\IEEEauthorrefmark{3}UC3M-Santander Big Data Institute}
    
}

        % <-this % stops a space
%\thanks{This paper was produced by the IEEE Publication Technology Group. They are in Piscataway, NJ.}% <-this % stops a space
%\thanks{Manuscript received April 19, 2021; revised August 16, 2021.}
%}



\maketitle

\begin{abstract}
The rapid expansion of Large Language Models (LLMs) has introduced unprecedented energy demands, extending beyond training to large-scale inference workloads that often dominate total lifecycle consumption. Deploying these models requires energy-intensive GPU infrastructure, and in some cases has even prompted plans to power data centers with nuclear energy. Despite this growing relevance, systematic analyses of inference energy consumption remain limited. In this work, we present a large-scale measurement-based study comprising over 32,500 measurements across 21 GPU configurations and 155 model architectures, from small open-source models to frontier systems. Using the vLLM inference engine, we quantify energy usage at the prompt level and identify how architectural and operational factors shape energy demand. Building on these insights, we develop a predictive model that accurately estimates inference energy consumption across unseen architectures and hardware, and implement it as a browser extension to raise awareness of the environmental impact of generative AI.
\end{abstract}


\input{1.Introduction}

\input{2.StateOfTheArt}
\input{3.Methodology}

\input{4.DataExploration}
\input{5.Model}

\input{6.PerformanceVsEnergy}
\input{7.ConsumptionPlugin}
\input{8.Conclusions}

\section*{Acknowledgments}

\input{Annex}

% {\appendix[Proof of the Zonklar Equations]
% Use $\backslash${\tt{appendix}} if you have a single appendix:
% Do not use $\backslash${\tt{section}} anymore after $\backslash${\tt{appendix}}, only $\backslash${\tt{section*}}.
% If you have multiple appendixes use $\backslash${\tt{appendices}} then use $\backslash${\tt{section}} to start each appendix.
% You must declare a $\backslash${\tt{section}} before using any $\backslash${\tt{subsection}} or using $\backslash${\tt{label}} ($\backslash${\tt{appendices}} by itself
%  starts a section numbered zero.)}


% \section{References Section}

\printbibliography

\vfill

\end{document}

%% file: 1.Introduction.tex
\section{Introduction}

Generative Artificial Intelligence (AI) has profoundly reshaped the landscape of modern technology and research, driving advances across science, industry, and everyday life. At the core of this transformation lies the Transformer architecture \cite{vaswani2017attention}, a foundational innovation that redefined how machines process and generate language. By introducing the concept of self-attention and enabling large-scale parallelization, Transformers overcame the limitations of earlier recurrent and convolutional models, unlocking new levels of scalability and efficiency. This breakthrough paved the way for the development of Large Language Models (LLMs), which have since achieved remarkable performance across an expanding range of tasks. Beyond natural language understanding and generation, LLMs now demonstrate strong capabilities in specialized domains such as code synthesis \cite{coignion2024performance}, text classification \cite{zhang2025pushing}, and mathematical reasoning \cite{guan2025rstarmathsmallllmsmaster}, illustrating their growing versatility and impact.

However, the rapid expansion of model size and deployment scale has introduced significant operational challenges. Beyond the computational cost of training and storage, practical concerns such as inference latency, model serving efficiency, and overall sustainability have become increasingly pressing for both AI researchers and industry practitioners. Among these issues, energy consumption has emerged as a particularly critical factor due to its direct environmental and economic implications. Large-scale AI facilities are highly dependent on stable and continuous power supplies, and the growing energy demands of modern LLM deployments have even sparked discussions of using nuclear energy. For instance, companies like Amazon and Google have announced plans to employ Small Modular Reactors (SMRs) to power portions of their data centers \cite{staffAmazonSignsAgreements2024,NewNuclearClean2024}. This trend highlights the unprecedented influence that LLMs now exert on global energy consumption.

While most discussions have traditionally focused on the energy required for model training, we argue that inference deserves greater attention. In practice, inference can account for up to 90\% of the total energy consumed over a model’s lifecycle \cite{AmazonEC2Update2019, walshHowMicrosoftMeasures2022, wuSustainableAIEnvironmental}. Understanding and accurately quantifying inference energy consumption is therefore essential. Existing research in this area remains limited: many studies examine only a narrow set of models, fail to consider variations in architectural characteristics, or overlook the differences between GPU types and multi-node deployments. As a result, substantial gaps persist in our understanding of the energy footprint of real-world LLM deployment. Addressing these gaps requires large-scale, systematic measurements across diverse hardware configurations and model architectures.

In this work, we present a comprehensive study comprising more than 32,500 measurements across nearly one million prompts. Our evaluation covers 21 distinct GPU configurations and 155 model architectures, including some of the largest open-source models such as Llama 3.1 405B and DeepSeek V3/R1 (685B), thus we cover an unprecedented range of models. All experiments were conducted in cloud-based environments using the state-of-the-art inference engine vLLM, thereby closely mirroring real-world LLM serving conditions. This setup enables us to measure the energy consumption of individual prompts with high granularity.

To this end, we adopt a fine-grained approach to analyzing how architectural factors affect energy usage. Specifically, we estimate the energy consumed per prompt, accounting for the fact that prompts are often processed in batches. This allows us to model consumption at the prompt level while also incorporating prompt-specific characteristics, such as input and output length, which strongly influence energy demand. In addition to these factors, energy consumption is also influenced by the underlying hardware and specific architectural details of the deployed model.

Finally, we introduce a predictive model that accurately estimates inference energy consumption across a wide range of architectures, including those not seen during training. Our results demonstrate that the model generalizes effectively, providing a practical tool for anticipating the energy requirements of future LLM deployments and supporting informed design and operational decisions.

As a practical demonstration, we implemented this model as a browser extension that estimates the energy usage of online LLM services such as ChatGPT, DeepSeek, and Gemini. The tool is publicly available, helping users understand the energy footprint of their interactions with AI systems and raising awareness about the environmental implications of generative AI. By bridging the gap between empirical measurement and predictive modeling, this study contributes not only to the growing field of sustainable AI but also to the broader effort of building scalable, responsible, and environmentally conscious AI infrastructure, raising awareness about the tangible environmental impact of everyday AI usage and encouraging more sustainable interaction with intelligent systems.

%% file: 2.StateOfTheArt.tex
\section{Related work}
\label{sec:headings}

A significant body of literature has been dedicated to analyzing the energy consumption of computing systems \cite{kor2015applications}, mobile devices \cite{balasubramanian2009energy}, and data center clusters \cite{aroca2015measurement}. Various software tools have been developed to measure energy consumption across these platforms, including Codecarbon \cite{benoit_courty_2024_11171501}, Carbontracker \cite{anthony2020carbontracker}, and Eco2AI \cite{eco2AI}. Some of these tools specifically target the measurement of GPU energy consumption in deep neural networks \cite{zeus-nsdi23}.

Large Language Models have gained substantial traction in natural language processing tasks; however, their inference and training incur considerable carbon costs. Training alone can require vast amounts of energy, for example, training the Llama 3.1 models consumed approximately 27.51 GWh, resulting in an estimated 11,390 tons of CO2 equivalent emissions \cite{MetallamaLlama31405BHugging2024}. 

In response to these environmental concerns, researchers have begun systematically investigating the carbon footprint of LLMs during both training and inference phases \cite{samsiWordsWattsBenchmarking2023d, luccioniEstimatingCarbonFootprint, pattersonCarbonEmissionsLarge2021b}. For instance, Nguyen et al. \cite{nguyenSustainableLargeLanguage2024} explored both operational and embodied emissions across LLaMA models (1B, 3B, and 7B) deployed on different GPUs, focusing on energy usage during the prefill and decode phases. Similarly, Patel et al. \cite{patelCharacterizingPowerManagement2024} analyzed power usage patterns during LLM training and inference in cloud environments. Their study spans a range of model sizes, from a few million parameters to 176B, and architectures, including encoder-decoder models like T5. They highlight the strong correlation between GPU peak power and cluster-wide peak consumption, emphasizing that GPUs are the dominant contributors to overall energy use in such systems.

Stojkovic et al. \cite{stojkovicGreenerLLMsBringing2024a} examined how different operational factors, such as tensor parallelism, GPU frequency scaling, batch sizes, and prompt characteristics, affect energy consumption. Adamska et al. \cite{adamskaGreenPrompting2025} focused on the inference energy cost of 7B-sized models, identifying output length as the most influential factor, and also investigated the impact of various prompt keywords on energy usage.

Chien et al. \cite{chienReducingCarbonImpact2023a} develop a workload model and predictive framework to assess the compute, energy, and carbon impacts of generative AI inference systems.  However, it is important to note that they do not analyse the inherent energy consumption characteristics of specific LLMs, but rather focus on the carbon footprint of the entire inference process by leveraging dynamic power grid carbon information and intelligent request direction algorithms to reduce emissions.
Wilkins et al. \cite{wilkinsOfflineEnergyOptimalLLM2024} also proposed a workload-based energy models tailored to LLM inference on heterogeneous systems. Although their method does involves profiling energy consumption and building accurate runtime and energy models for various LLMs. Although they do not create a generalistic model, rather a regression model for each LLM. In contrast, Faiz et al. \cite{faizLLMCARBONMODELINGENDTOEND2024a} introduced an end-to-end carbon modeling framework for both training and inference. However, their inference experiments were constrained to a single GPU configuration, fixed batch size, and a constant number of input tokens, limiting their applicability to more dynamic real-world settings. Addressing these limitations, Fu et al. \cite{fuLLMCO2AdvancingAccurate2024} proposed LLMCO2, a graph neural network (GNN)-based model designed to predict the carbon footprint of LLM inference more accurately. Their approach considers variables such as prompt length and hardware specifications. Nonetheless, their dataset focuses on inference scenarios with batch sizes under two.

There are some commercial solutions, such as the Carbon Footprint Calculator \cite{carbon_footprint_calculator}. However, these tools are limited, as they do not account for batch size in their calculations, and they assign the same carbon footprint to different models, even when the models vary substantially in size but appear similar.

\begin{table*}[tbp]
\caption{Comparison with other studies}
\label{tab:comparison}
\scriptsize
\begin{tabular}{lcccccccc}
\hline
Paper & Pred. model & HW Features & Batch size & +30 LLMs & 100B+ LLMs & Different GPUs & Multi-node GPU & Impact LLM Structure \\ \hline
Samsi et al. \cite{samsiWordsWattsBenchmarking2023d} &  & \checkmark & \checkmark &  &  & \checkmark &  &  \\
Luccioni et al. \cite{luccioniEstimatingCarbonFootprint} &  & \checkmark & \checkmark &  & \checkmark & \checkmark & \checkmark &  \\
Patterson et al. \cite{pattersonCarbonEmissionsLarge2021b} &  & \checkmark &  &  & \checkmark & \checkmark & \checkmark &  \\
Nguyen et al. \cite{nguyenSustainableLargeLanguage2024} &  &  & \checkmark &  &  & \checkmark &  &  \\
Patel et al. \cite{patelCharacterizingPowerManagement2024} &  &  & \checkmark &  & \checkmark &  & \checkmark &  \\
Stojkovic et al. \cite{stojkovicGreenerLLMsBringing2024a} &  & \checkmark & \checkmark &  &  &  &  &  \\
Adamska et al. \cite{adamskaGreenPrompting2025} &  &  &  &  &  &  &  &  \\
Chien et al. \cite{chienReducingCarbonImpact2023a} & \checkmark(*) & \checkmark &  &  & \checkmark & \checkmark & \checkmark &  \\
Wilkins et al. \cite{wilkinsOfflineEnergyOptimalLLM2024} & \checkmark &  &  &  &  &  &  &  \\
Faiz et al. \cite{faizLLMCARBONMODELINGENDTOEND2024a} & \checkmark &  &  &  &  &  &  &  \\
Fu et al. \cite{fuLLMCO2AdvancingAccurate2024} & \checkmark & \checkmark & \checkmark(**) &  &  & \checkmark &  &  \\
\hline
Ours & \checkmark &  \checkmark & \checkmark & \checkmark & \checkmark & \checkmark & \checkmark & \checkmark \\

\hline
\end{tabular}
\vspace{2mm}
\vfill
\parbox{0.8\linewidth}{
\footnotesize
(*) Request direction optimization model.
\vfill
(**) Their model allows batch size but their data collection focuses on batches sizes under two.
}
\end{table*}

Table \ref{tab:comparison} presents a summary of existing academic studies that analyze the energy consumption of large language models, highlighting their key features and comparing them with our work. A common limitation among these studies is their narrow focus on a small number of models, often limited to a single family. This poses a challenge, as different architectures can lead to significantly different energy usage patterns. Additionally, to the best of our knowledge, previous research has not explored how specific architectural aspects of LLMs, such as the number of layers, affect energy consumption, which limits the ability to generalize findings or build predictive models. Our approach supports improved generalization and enables extrapolation to newer or unseen models.

%% file: 3.Methodology.tex
\section{Methodology}

This paper aims to understand what defines the energy cost of an LLM inference, given that in a production setting, the cluster will attend to several concurrent prompts simultaneously.  In particular, we will focus on the decoder-only (or causal decoder) architecture \cite{naveedComprehensiveOverviewLarge2024}, as this is the most common architecture in newly trained models, as they are easier to train unsupervisedly, as well as having better throughput than encoder-decoder models.

One key element of this study is that we focus on power consumption, not the inferring process's speed performance. Although both of them are very related, i.e., a faster LLM with the same accelerator will require less energy than a slower one, it is essential to consider that there are some scenarios in which, for example, a GPU will not be working to its maximum capacity (i.e., not reaching their max TDP). Therefore, our focus in this study is not on how fast a particular model is (e.g., token speed or Time to First Token) but more on how energy-efficient the different models we have studied are. As there is a limited number of LLMs, we will also modify the structure of some of them to see better the effect of changing some characteristics, such as the number of layers or their dimensionality.

Thus, we will test a wide range of scenarios of prompts of the same size (i.e., with number of input and output tokens), measuring that energy consumption for a particular GPU configuration (i.e., the model and the number of GPU). Knowing the batch size, we can derive the energy cost of a single prompt. We assume this particular scenario, as the number of operations is the same for each of the prompts, i.e., all prompts go through the LLM layers the same amount of time, generates the same amounts of tokens.

Therefore, our goal is to gather as many measurements as possible in order to build a model that estimates energy consumption based on the structure of a model and the prompts with their responses. This task is not straightforward, as it involves several challenges. For example, not all models can be used with every GPU configuration. Naturally, if a model requires more memory than what is available on a given GPU, we cannot load that model (we also do not use CPU offloading). In cases where a model exceeds the maximum memory capacity of a single server node, such as with models like \texttt{deepseek-ai/DeepSeek-V3} or \texttt{meta-llama/Llama-3.1-405B}, we deploy multiple servers in order to leverage multi-node configurations.

For each of the models, we gather some information about their structure. This includes a different range of parameters: model size (in number of parameters), embedding, self-attention, FFN parameters, number of layers, dimensionality of the model, attention heads, mechanism used for self-attention, or whether the model is a Mixture of Experts (MoE).

For the GPUs, we also collect different metrics: GPU memory, TFLOPS (depending on the precision of the model), CUDA Cores, GPU Bandwidth, TDP. But also other aspects, such as the amount of memory free once the model has been loaded, or the size that a prompt can take up in the KV cache. 

It is also relevant to consider that when using several GPUs, we are not considering Data Parallelism but rather Model parallelism.

\subsection{Inference Engine}

The models are being run using the vLLM library \cite{kwon2023efficient}, a state-of-the-art inference engine which is widely used for its memory management that leverages high speed token generation. We decided to use vLLM to  measure energy consumption as: (i) it is widely used in production settings; (ii) it provides support for a great number of model archiectures; (iii) it provides PagedAttention which optimizes inferences as it manages better memory. Furthermore, there are some models which even recommended to be deployed using vLLM (e.g., Qwen2.5-72B \cite{qwen2.5}, DeepSeek-V3 \cite{deepseek-aideepseek-v3_2025}). 

There is also the advantage of allowing continuous batching, i.e., the prefill stage can happen while the decoding phase of previous prompts takes place. This allows faster processing, as the engine attends to a different number of requests simultaneously, which can vary over time or as the prompts are processed, this functionality is key in high-load scenarios. 

However, there are other libraries such as SGLang \cite{zheng2024sglang} or TensorRT-LLM \cite{NVIDIATensorRTLLM2025}, which can perform at the same level or faster than vLLM for some models but under specific scenarios. 

\subsection{Energy Measurement}

For our measurements, we focus exclusively on the GPU's energy impact for two main reasons. First, the GPU is by far the most energy-demanding component during inference. A previous study showed that the CPU, RAM, and disk together account for only about 12\% of total energy consumption \cite{10549890}. Second, we conduct our measurements using virtual machine instances on Google Cloud. Running on these virtualized servers allows us to evaluate different GPUs cost-effectively. However, this setup comes with some limitations: (i) we do not have access to the physical hardware, which prevents us from directly measuring components such as RAM or disks; and (ii) RAPL files are not available on virtualized servers, so we cannot obtain accurate CPU energy measurements using this method either.

We carry out the measurements using Codecarbon \cite{benoit_courty_2024_11171501}, which uses the NVIDIA Management Library (NVML). This library provides an interface for monitoring some features of NVIDIA GPUs, in particular Power Draw. We will periodically sample the power used by the GPUs, to then have an energy measurement. 

These measurements focus on the total energy consumed throughout the entire inference process. As such, we do not analyze token generation speed or break down the energy consumption by individual phases of inference. Specifically, we do not measure the energy used during the prefill and decode phases separately. Our reasoning is based on the continuous batching mechanism used by vLLM, where a query can enter the prefill stage while another is in the decode stage. However, by designing specific scenarios, we can still gain insight into each phase: when using inputs with many tokens and few outputs, we primarily capture the energy usage of the prefill stage; conversely, when using minimal input and generating many output tokens, we effectively capture the energy consumption of the decode phase.

\subsection{LLM Selection}

In this paper, we aimed for a comprehensive selection of LLMs to support the development of a generalist model that performs reliably across diverse architectures. We focused on state-of-the-art open-source models, prioritizing popularity and coverage across different model sizes. This process resulted in a set of 55 root LLMs, listed in Annex \ref{annex:models_selected}.

To further investigate how specific architectural factors, such as the number of layers, affect energy consumption, we created additional model variants by systematically modifying one characteristic at a time. This allowed us to (i) analyze the impact of individual features on energy usage and (ii) refine our estimation model to better capture architectural effects. Through this procedure, the total number of evaluated models increased to 155.

\subsection{Experiment design}

For each selected LLM, we performed measurements across a grid of input and output token sizes, enabling us to capture their interaction effects. These measurements were conducted for both single prompts and batched prompts, with all prompts in a batch configured to the same input and output lengths to ensure accurate per-prompt comparisons.

The grid was limited to token values below 1,000, following Perez-Ramirez et al. \cite{perezramirez2025castillocharacterizingresponselength}, who showed that publicly available datasets of open-ended instructions, problem-solving tasks, and code generation rarely exceed 1,000 tokens—even at the 99th percentile. Similar patterns were observed for average response lengths. We therefore consider this threshold sufficient to capture the majority of common LLM usage scenarios.

This setup does not cover all possible use cases. To address this, we conducted additional inference experiments using a broader range of batch sizes with a reduced token grid. Since some applications require longer context windows, we also included experiments with input lengths extending to tens of thousands of tokens, allowing us to capture such scenarios more accurately. These complementary tests provide a broader view of LLM behavior.

We further repeated measurements across different GPU configurations, specifically NVIDIA V100, T4, L4, A100, and H100. These GPUs vary in capabilities, memory, bandwidth, and computational power (TFLOPs). Incorporating this diversity enables us to model LLM energy consumption while accounting for the impact of underlying hardware.

\subsection{Elements to Evaluate}

Based on our experimental design, we aim to examine several features and factors that may influence energy consumption.  

\begin{enumerate}
    \item Content length: Includes both input and output tokens. The amount of text processed or generated can have a significant impact.
    \item Batch size: Refers to the number of samples processed simultaneously, which can affect efficiency and resource usage.
    \item Architectural details: Covers model size as well as more specific aspects such as the number of layers, hidden dimensions, and other structural properties.
    \item Quantized models: Models that use reduced precision representations to lower computational demands.
    \item Hardware differences: Considers the impact of using different types of GPUs, as well as the effect of scaling the number of devices.
\end{enumerate}

%% file: 4.DataExploration.tex
\section{Data Exploration and variables explainability}

In this section, we explore some of the metrics of the dataset we have collected, which already gives an idea on how the architecture of these models and the selected hardware defines how these models consume energy. 

We have made more than {32,500} different measurements which accounts for almost one million prompts, using {21} different GPU configurations, with a total of {149} different models.

\subsection{Content length: Input and Output Tokens}

Due to the transformers' nature, it is clear that the amount of tokens generated will determine the energy consumed, i.e., the inferring process is repeated for each token. However, also having longer prompts fed into the transformer means that bigger matrix operations need to be performed. 

\begin{figure}[ht]
    \centering
    \includegraphics[width=.9\linewidth]{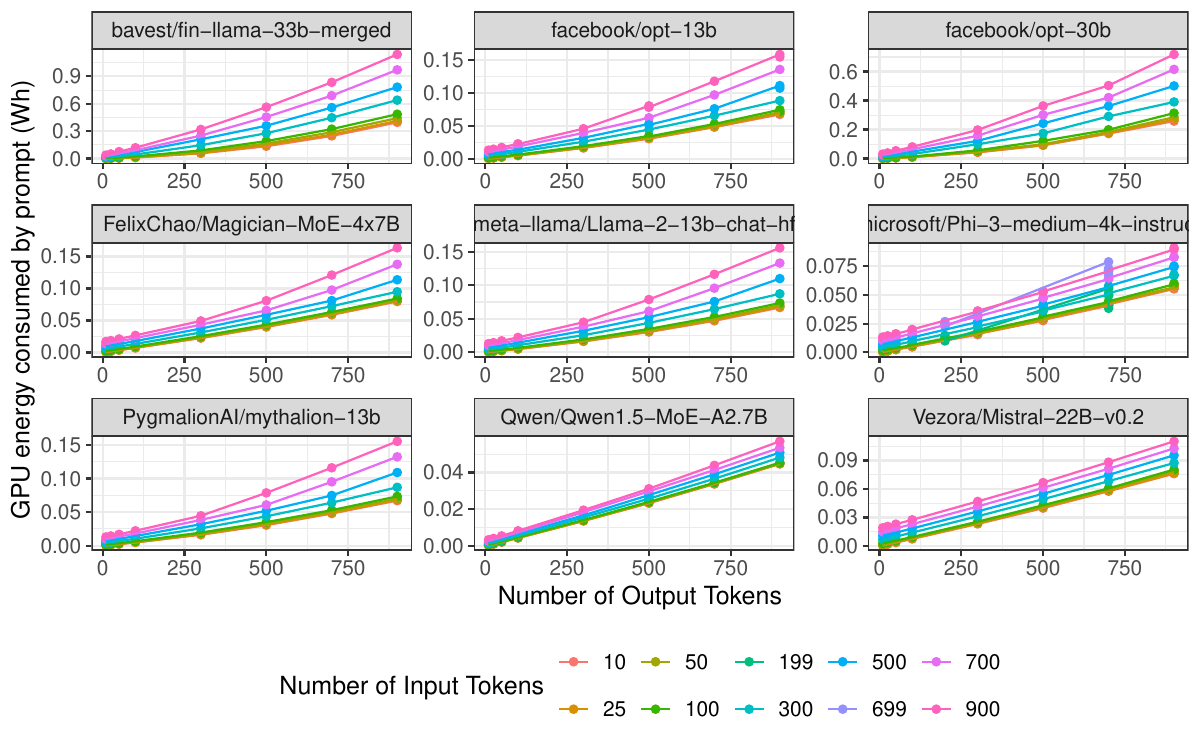}
    \caption{The effect of increasing the amount of input and output tokens in different models using an NVIDIA A100 80GB as accelerator.}
    \label{fig:input_output_matrixes}
\end{figure}

In Figure \ref{fig:input_output_matrixes}, we show how input and output tokens affect the GPU’s energy consumption per prompt across nine models using an NVIDIA A100 80GB accelerator. The results reveal two key points: (i) both input tokens (\TokIn) and output tokens (\TokOut) contribute to energy consumption in a linear or near-linear manner, and (ii) output tokens have a much stronger impact than input tokens. Although the figure presents only nine models, the same patterns are observed across all other models we evaluated.

Using the model \texttt{facebook/opt-30b} as a reference, we observe that a prompt $P$ with 100 input and 100 output tokens, denoted as $P_{1}$(100\TokIn, 100\TokOut), consumes approximately 0.0137 Wh of energy. When the prompt length increases to 900 input tokens while maintaining 100 output tokens ($P_{2}$(900\TokIn, 100\TokOut)), the energy consumption rises to 0.08 Wh. Conversely, with a short input and a much longer output ($P_{3}$(100\TokIn, 900\TokOut)), the consumption reaches 0.3 Wh, more than three times that of $P_{2}$.

To provide some context for these prompts: $P_{1}$ could represent a short translation task, $P_{2}$ might correspond to a summarization task, and $P_{3}$ could reflect a generative task, such as open-ended text generation or question answering.

Across all models using this GPU configuration, $P_{2}$ prompts consume 2.19 times more energy than $P_{1}$, while $P_{3}$ prompts consume 11.00 times more. These results demonstrate that the number of output tokens has a significantly greater impact on the energy consumption of large language models compared to the number of input tokens.

\subsubsection{Long context}

Long context capabilities are crucial for diverse applications, ranging from analyzing extensive documents to engaging in extended chatbot conversations. However, managing long context presents significant challenges. Computationally, methods like sliding windows exist, but they often compromise performance. Memory management is also a key concern, particularly regarding the Key-Value (KV) cache. Longer prompts demand more KV cache memory, which can strain systems with limited resources and increase energy consumption.

\begin{figure}
    \centering
    \includegraphics[width=\linewidth]{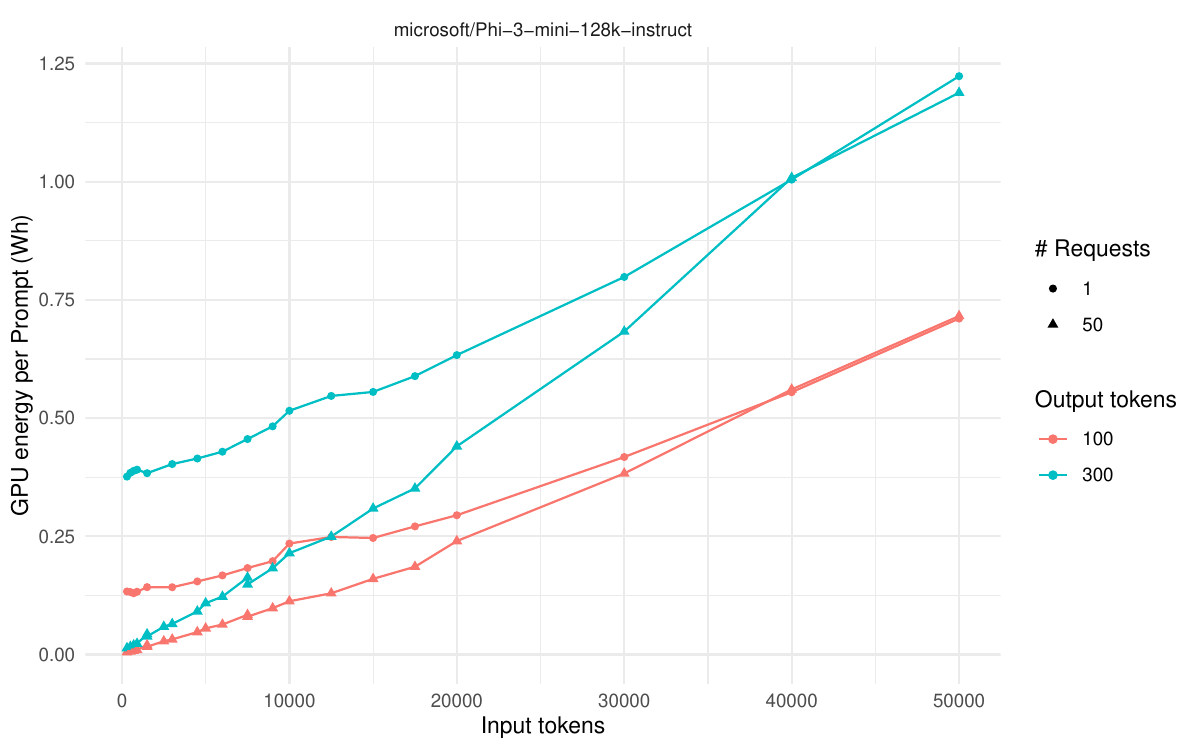}
    \caption{Energy consumption with long context. Measurements done with 2 NVIDIA L4 GPUs with the Phi-3 mini 128k context size.}
    \label{fig:long_context}
\end{figure}

To illustrate this, we measured the energy consumption of the Phi-3 mini model with a 128k context window, deployed on a system with two NVIDIA L4 GPUs (Figure \ref{fig:long_context}). After loading the model, we observed that we could effectively work with context lengths exceeding 50k tokens, the limit imposed by the KV cache capacity. This memory demand directly impacts the number of concurrent prompts the system can handle, ultimately increasing energy consumption.

A key observation is that once the context length reaches the KV cache's capacity, the energy consumption for processing a single request becomes nearly indistinguishable from that of processing  multiple requests.

However, processing very long context inputs with only one request at a time does not lead to as drastic an increase in energy consumption as one might anticipate. For example, a prompt with 50,000 input tokens consumes only 3.25 times more energy than a prompt with 300 input tokens. 

The most significant conclusion is that it becomes crucial to consider whether the KV cache will be fully utilized. Ultimately, the primary bottleneck for LLM inference in current systems is available memory.

\subsection{Batch size}

Modern hardware accelerators are designed to efficiently process multiple prompts in parallel, significantly reducing energy consumption. When a system handles several prompts at once, the total inference time can remain nearly the same as processing a single prompt. For instance, if one prompt takes as long as a batch of ten, the energy used per prompt is effectively reduced by a factor of ten.

The continuous batching capability of vLLM enables dynamic adjustment of batch sizes based on prompt characteristics. More importantly, this feature increases throughput and lowers energy usage per generation. However, in real-world scenarios, latency is often prioritized. As a result, systems may operate below their maximum batch size potential, limiting the efficiency gains typically seen in offline generation.

\begin{figure}[ht]
\centering
\includegraphics[width=\linewidth]{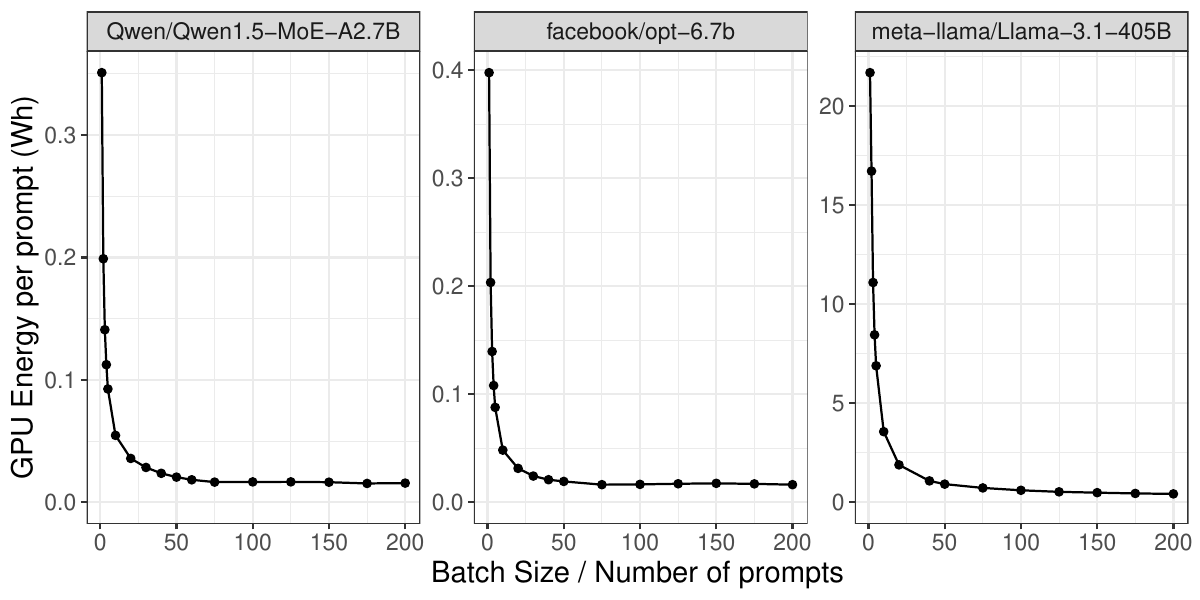}
\caption{\textbf{Energy consumption by number of prompts}. The plot shows how GPU energy per prompt decreases with increasing batch size. Each prompt contains 300 \TokIn and 300 \TokOut.}
\label{fig:energy_prompts}
\end{figure}

Figure~\ref{fig:energy_prompts} shows how energy consumption per prompt varies as the number of prompts increases, across different models and accelerators. Note that the effective batch size used by vLLM depends on the characteristics of the prompts, so the actual number of processed prompts may differ based on how the vLLM engine allocates resources. In this figure, we report the number of prompts supplied simultaneously.

All scenarios use prompts and responses of equal size $P$(300\TokIn, 300\TokOut). The models \texttt{Qwen/Qwen1.5-MoE-A2.7B} and \texttt{facebook/opt-6.7b} were tested using two NVIDIA L4 GPUs, while \texttt{meta-llama/Llama-3.1-405B} ran on a cluster of 2 x 8 NVIDIA A100 80GB GPUs. As shown in the figure, using a single prompt per batch is inefficient both in energy usage and performance. For instance, the Llama 405B model takes 15.4 seconds and consumes 21.7Wh for a single prompt. In contrast, processing a batch of 100 prompts takes only 29.4 seconds and consumes 60.4Wh, which results in significantly lower energy consumption per prompt. This pattern is present for all the studied LLMs.

This demonstrates that maximizing batch size not only improves throughput but also enables large-scale models to operate more sustainably and cost-effectively in real-world applications.

\subsection{LLMs' Architectural Details}

Typically, the model size or number of parameters is widely used to define how ``big'' a LLM is. With these models we see models which varies from the hundreds millions of parameters (e.g., \texttt{facebook/opt-125m} with 125.2M parameters) to the few hundreds of billions (e.g., \texttt{meta-llama/Llama-3.1-405B} with 405.8B parameters). Although this size may indicate the accelerators needed (in terms of memory), to run these models, it does not necessarily explains the amount of time of execution (which can be an indicative of the power consumption). 

At the end these parameters, may be encoded in different parts of the transformer, e.g., two different models may have the same size, but the distribution of these differs. For instance, \texttt{facebook/opt-2.7b} and \texttt{google/gemma-2b} both have a similar number of parameters, 2.65B and 2.51B respectively. Whereas the former has more layers (32) than the Gemma model (18). But in contraposition the Gemma model has 110M parameters per layers against the 79M of the OPT, with 4 time more embedding parameters (524M against the 133M). And more remarkable, although the both of them have the same size, this is not translated in the same energy consumption, for a $P$(500\TokIn, 500\TokOut), the OTP model consumes 18.5mW, whereas the Gemma consumes 8.6mW (more than twice the energy), both with the same number prompts and accelerator (1 NVIDIA L4). 

There are also models that use Mixture of Experts (MoE), which does not activate all of its weights during the inference of a single token, therefore although they are bigger models their inference may be equivalent to the one of a smaller model. 

\subsubsection{Model Size}

\begin{figure}[ht]
    \centering
    \includegraphics[width=1\linewidth]{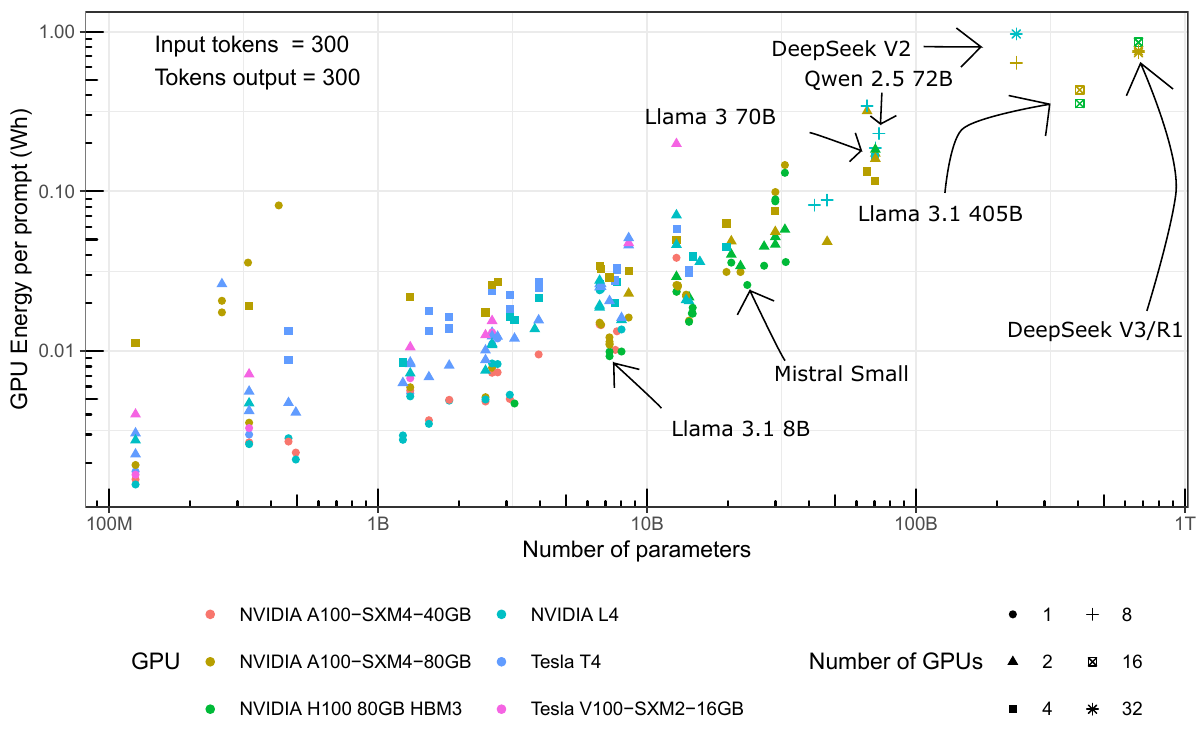}
    \caption{\textbf{Model Parameters against energy consumed.} This plot shows the energy consumed using LLMs by prompt using prompts with 500 input and output tokens.}
    \label{fig:parameters_size}
\end{figure}

Figure \ref{fig:parameters_size} shows the energy consumed by prompt per model according to their model size. As it can be observed, the number of parameters is not a direct proxy to energy consumption, while it clearly gives the idea of bigger models consuming more, there are more nuances to consider in a model. 

If we fit a linear regressor using only the number of parameters in the LLM, we obtain an $R^2$ of 0.49. This suggests that model size alone does not fully account for energy consumption, as models with the same number of parameters can differ significantly: some consuming nearly an order of magnitude more than others. That said, as a general rule of thumb, larger models do tend to consume several orders of magnitude more energy than smaller ones.

\subsubsection{Number of Layers}

Transformers are composed of a series of consecutive layers, in which each of them relies on the result of the previous layer. Therefore to process a single token, it has to "travel through" all layers of the LLM, thus meaning that increasing the size of layers will definitely impact the energy consumed. 

The main load while generating a new token is found in the layers of the decoder. Although the design or architecture of these layers may vary from LLM to LLM, they should remain the same within the model, although there are some novel mechanism such as layer skip or early exiting \cite{he2025adaskip, chen2023ee}, although they are not widely implemented. This means that all layers will perform the same number operations, therefore is expected that the energy of a model with $N/2$ layers will be the half of a model with $N$ layers.

\begin{figure}[ht]
    \centering
    \includegraphics[width=1\linewidth]{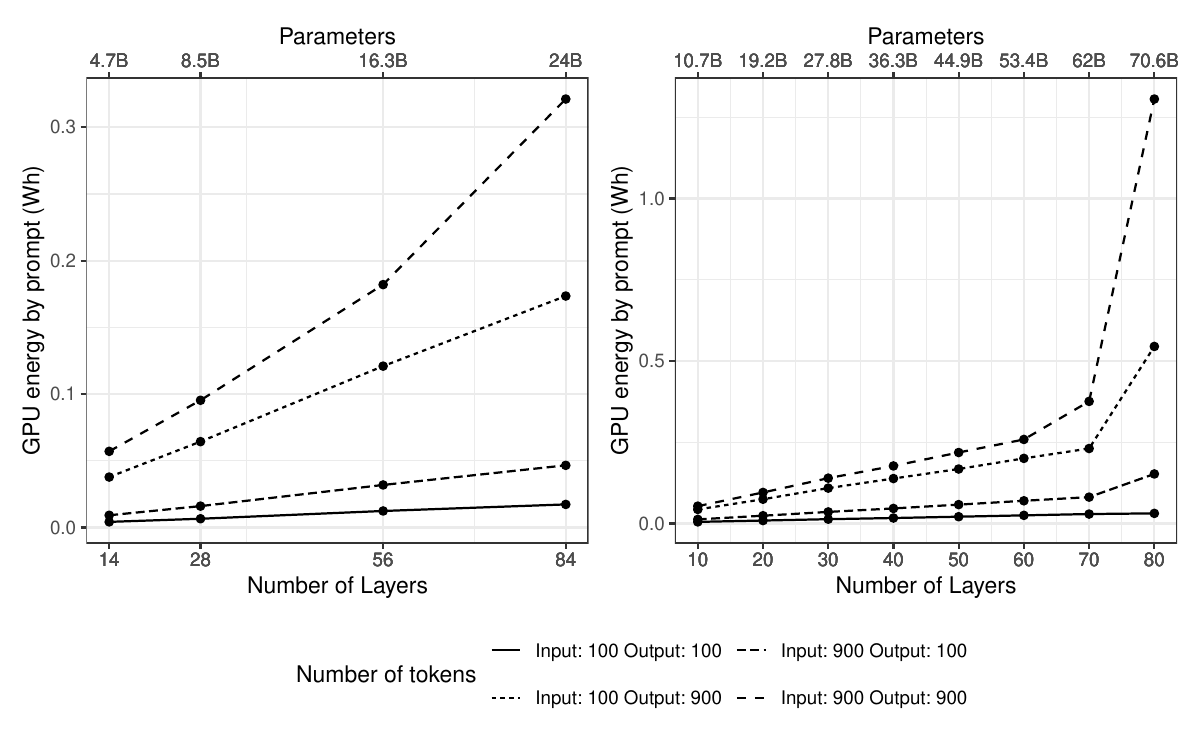}
    \caption{\textbf{Energy consumption by number of layers in the Gemma 7B model}. We selected two models and modified the amount of layers, these experiments were run in 2 NVIDIA A100 80GB running the models with BF16.}
    \label{fig:layers_inscrease}
\end{figure}

We have observed the expected linearity (with some caveats) with both Gemma and Llama 3 models, as seen in Figure \ref{fig:layers_inscrease}. Specifically we used the Gemma 7B and the Llama 3 70B models for this experiment, these models originally have 28 and 80 layers, the rest of the values remains unchanged. For the Gemma Model, we explored with up to 84 layers. Whereas is the case of the Llama 3 70B, we explored lower values. And for the most part, the layer effect remains linear as expected, with the exception of the original Llama 3 models. The reason this happens is by the low amount of VRAM free for the KV cache, in particular as we used 2 A100 80GB, having the default values in vLLM (gpu usage is 0.9), the majority of the memory will be used by the model weights. The KV cache, therefore only have a few GB allocated, in particular in this scenario it can only allocated around 18,000 tokens, which is not enough for all the requests.

Obviously, the design of the layer will heavily impact the consumption of the LLM, therefore in the following subsections, we will dive deeper into some of the aspects to consider with different models.

\subsubsection{Dimensionality of the models}

While referring to the dimensionality of the LLMs, we may refer to two different aspects of one layer. On the one hand we have the model dimension: $d_{model}$\footnote{As used in \cite{vaswani2017attention}} or \texttt{hidden\_size}\footnote{As typically used in model cards in HuggingFace}, which is the dimension of the embedding and the input and outputs of the layers. This dimension is key for performance, as from it depends the operations for the Self-Attention blocks and the MLP. But for the MLP or FFN it is also necessary to consider the intermediate size, which is the dimensionality inside the FFN blocks. This is usually bigger than the hidden size.

In here we have to consider a lot of things, even the vocab size, the hidden size, size of the MLP.

\begin{figure}
    \centering
    \includegraphics[width=\linewidth]{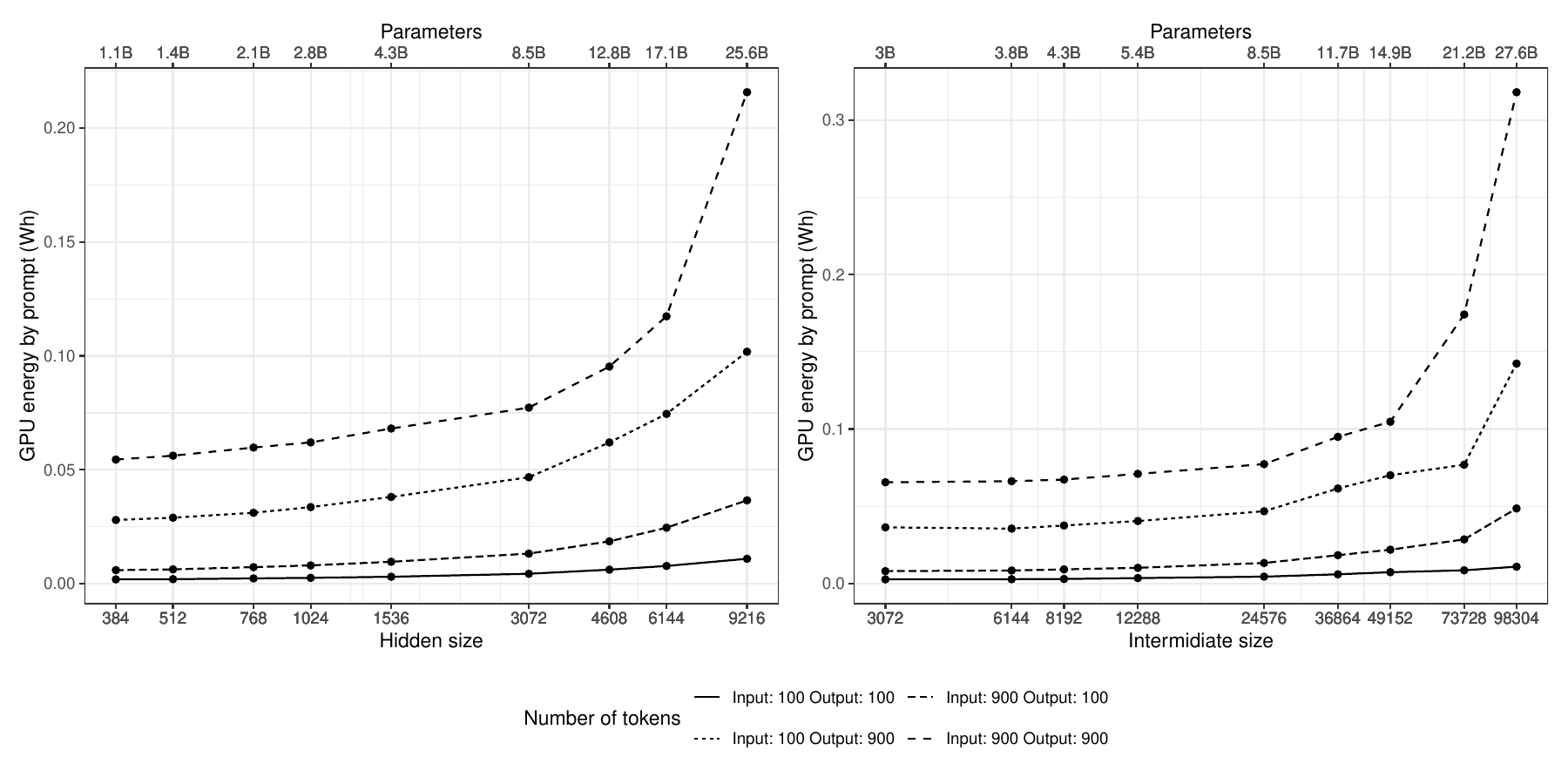}
    \caption{Energy consumption by the hidden size and the intermediate size in the Gemma 7B model}
    \label{fig:hidden_size}
\end{figure}

In this case we have experimented with the Gemma 7B and a A100 80GB, with different hidden sizes, we used values derived from the original value (3072), all of other parameters of the model remain the same. This experiment can be seen in the left side of Figure \ref{fig:hidden_size}. With lower values (from 384 to 1536), the energy consumed remains relatively similar: however, this modification in the hidden size obliviously reduces the number of parameters of the model, for instance, the smallest model (i.e., 384 of hidden size and 1.1B parameters), consumes almost the same energy than the one with 2.8B: this is a clear indication that relying only on the parameter count is not feasible to understand the energy consumption of a certain model. Although this is out of the scope of this paper, it is reasonable to acknowledge that reducing this hidden size so much (e.g., 384), will not be very productive, as the energy impact will be almost the same, but one could expect that the quality of the responses to be much worse. 

Regarding the Feed Forward Network (FNN), typically it has a higher dimensionality than the hidden size due to the use of intermediate projection layers that expand and contract the representation space, which enhances the model’s capacity to learn complex patterns \cite{vaswani2017attention}. This FNN block is usually have an up projection (increasing the dimensionality from the hidden size), and the down projection (i.e., returning the hidden size dimensionality). For instance, in the Gemma 7B, the hidden size is 3072, but the intermediate size is 24576 (which is exactly 8 times more).

These features and the number of layers of a model are critical architectural features that significantly influence the model’s performance in generating high-quality responses \cite{wuPerformanceLawLarge2024}.

\subsubsection{Self-Attention Mechanism}

The attention in LLM is an essential part of the transformer architecture, which allows the model to focus on the relevant parts of a sequence. Although how this attention is computed can vary from LLM to LLM. For instance, one thing to consider is how the attention is computed. Generally, attention is divided into multiple heads to be able to focus on different parts of the input, therefore having an LLM with very few heads will not allow the model to fully grasp the text, for all models we have made measurements the median value are 32 different attention heads. 

There are multiple ways to implement the self-attention mechanism, for instance, the base case of Multi Head Attention, where each head computes its own query, value, and key. But there is also the case where all heads share the same value and key (Multi Query Attention or MQA) \cite{shazeer2019fast}, which improves the inferring speed but sacrifices some of the performance of the model. In between, there is the Group Query Attention (GQA) \cite{ainslie2023gqa}, in each a key value pair is shared by several heads. 

Obviously it is clear that MHA is the option that requires more time to process, therefore it will consume the most power. The consumed energy will also depend on the number of heads

\begin{figure}
    \centering
    \includegraphics[width=1\linewidth]{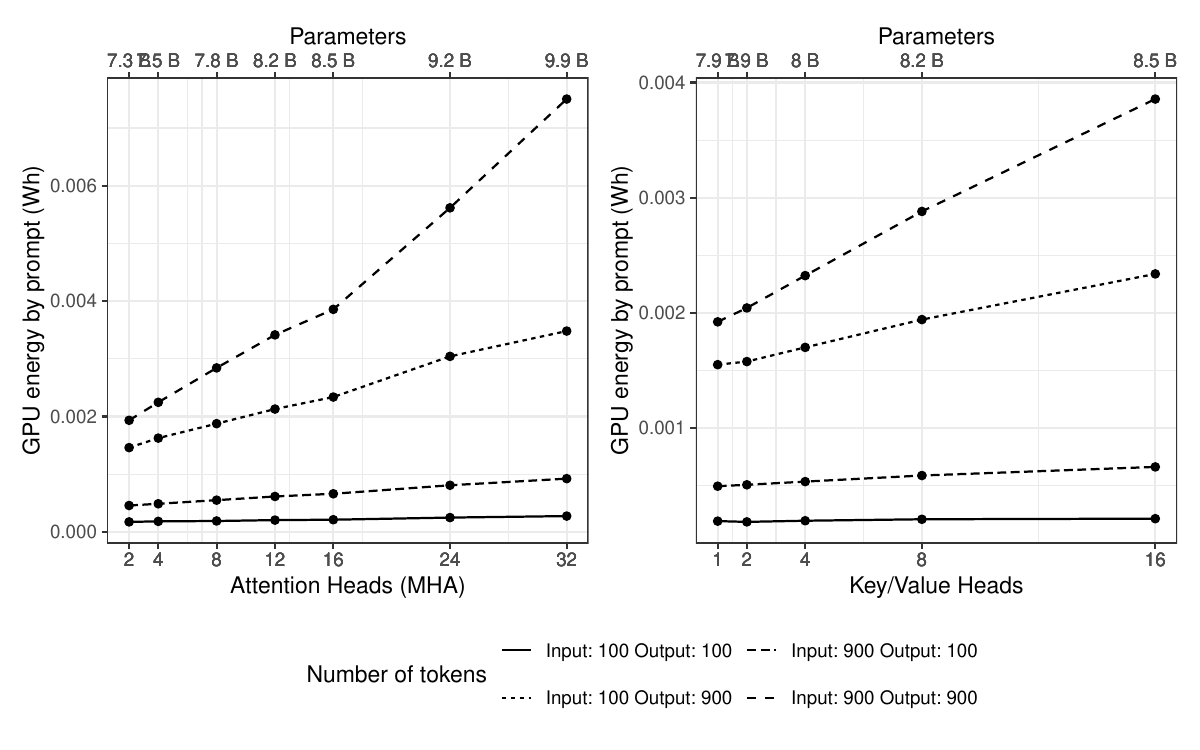}
    \caption{Energy consumption by modifying attributes of the Self-Attention mechanism of Gemma 7B model. Left side chart shows the variation of Attention heads (original model has 16). Right side shows the variation of KV Heads, showing MQA, GQA and MHA energy usage.}
    \label{fig:self_attention}
\end{figure}

As seen in figure \ref{fig:self_attention}, we have modified the Gemma 7B model, which by default uses MHA and has 16 Attention Heads (and therefore 16 key/value heads). On the left side, we depict the change of attention heads (and parameters) of such model with the MHA mechanism. Whereas on the right side of the chart, we see the variation of KV Heads with 16 Attention Heads, therefore we can see MHA (16 KV Heads), MQA (only one key value head), and GQA (the intermediate values). MQA provides the best energy performance, but GQA seems to have a better trafe-off between energy and model performance, and it is a more preferred approach than both MHA and MQA. 

%\subsubsection{Mixture of Experts}

%MoE allows sparse execution of models, therefore they do not activate all their parameters in runtime. This means that the computation is faster, although they still have the drawback that all the model should be loaded, therefore a 8x7B model, although it is faster than a 56B models, they still need to allocate the same amount of vRAM towards this task. Therefore, the GPU memory is still one of the major bottlenecks, as if there is not enough space to handle too many requests, the KV cache is going to be filled up and therefore the inference will became expensive. Therefore these models are not recommended when there is a lack of VRAM for that same reason. 

\subsection{Quantified models: different dtypes}

Model precision is one of the most critical factors influencing the performance of transformer-based computations. This is primarily due to two reasons: (i) hardware accelerators achieve significantly higher throughput with lower-precision data types (for example, the NVIDIA L4 delivers 120 TFLOPS for FP32, 242 TFLOPS for FP16, and 485 TFLOPS for FP8); and (ii) lower-precision formats require less memory, enabling more efficient utilization of hardware resources.

Quantization is a widely used technique that reduces the precision of model parameters to improve inference speed and reduce memory usage. By lowering memory requirements, quantization enables large models to run on low-resource systems. For instance, a 13B model in FP16 (such as OPT-13B) cannot fit into a single NVIDIA L4 GPU, but a quantized version of the same model can be loaded and executed successfully.

Common quantization methods include AWQ \cite{lin2024awq}, GGUF, BNB \cite{dettmers2023qlora}, and GPTQ \cite{frantar2022gptq}.

\begin{figure}
\centering
\includegraphics[width=.8\linewidth]{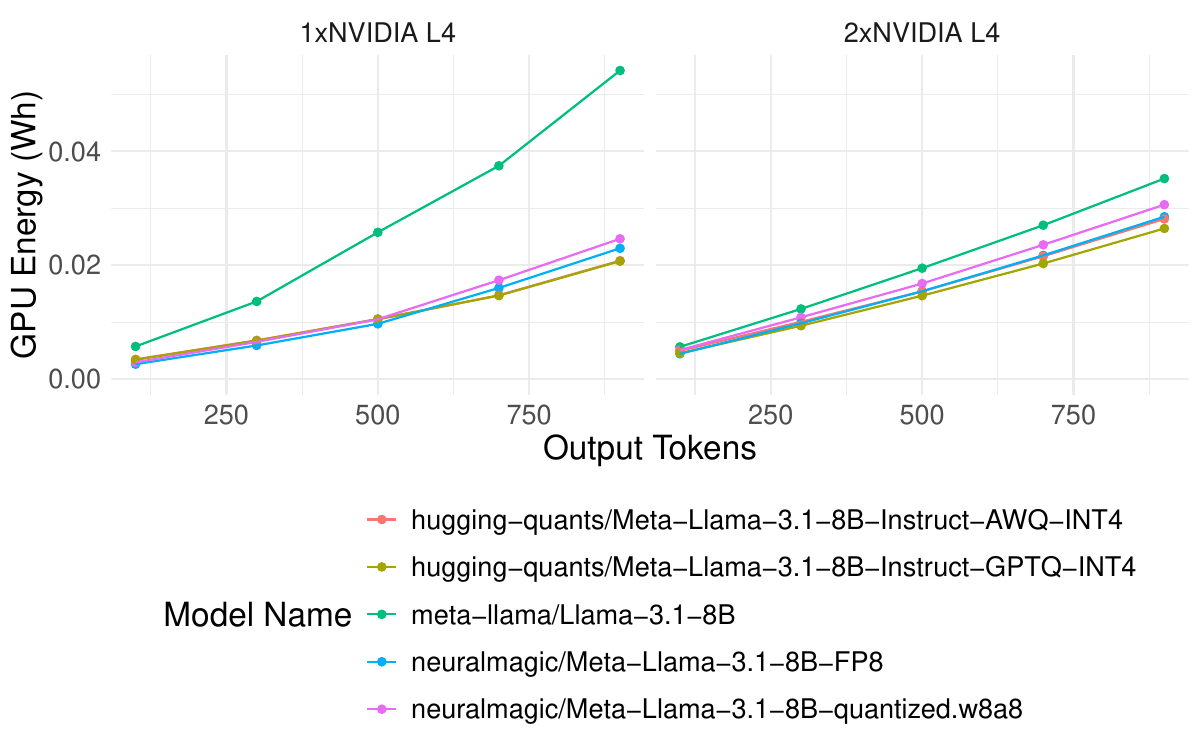}
\caption{Quantization Experiment}
\label{fig:quantization}
\end{figure}

Figure \ref{fig:quantization} shows the energy consumption of various quantized versions of the LLaMA 3.1 8B model under two configurations: a single NVIDIA L4 and two NVIDIA L4s.

As shown in the figure, the unquantized model consumes the most energy. This is especially evident when using a single NVIDIA L4, where energy consumption is significantly higher compared to quantized versions. This can be attributed to the memory overhead required to load the full 8B model onto a single GPU, which pushes the hardware beyond its optimal operating range. In contrast, when two NVIDIA L4s are available, the system has sufficient memory for the model and the KV cache. In this case, the energy savings provided by quantization become less substantial, while potential losses in model quality due to reduced precision may outweigh the benefits.

Thus, quantization proves to be an effective strategy for reducing energy consumption in resource-constrained environments. However, it is important to recognize the trade-off between energy efficiency and output quality, which depends on the quantization method used. In scenarios where memory is not a bottleneck, the benefits of quantization may be minimal and not justify the potential degradation in performance.

% \begin{figure}
%     \centering
%     \includegraphics[width=.8\linewidth]{plots/16_and_32_precision.pdf}
%     \caption{OPT 1.3B, Gemma 2B and Llama 3.1 8B models running with precision 16 and 32}
%     \label{fig:16_32_precision}
% \end{figure}

%Another point of view is having a data type with more precision than in the original model. The Llama3.1 8B model was designed with BF16, but it could technically be executed under FP32 precision. In Figure \ref{fig:16_32_precision}, we run another experiment comparing 16 and 32-bit precision. If the Llama 3.1 runs at higher precision, we will face the same issue as running the BF16 precision model with a single NVIDIA L4: low memory available for the KV cache, which will slow down inference, increasing the energy consumption. 

\subsection{Hardware Effect}

As previously mentioned, we conducted experiments using a range of GPU configurations, from a single NVIDIA T4 to clusters with up to 4×8 A100 GPUs, to run large models such as LLaMA 3 405B and DeepSeek V3. Each hardware setup has a distinct energy usage profile, shaped by factors such as GPU performance, memory bandwidth, and thermal design power (TDP). Since these parameters vary depending on the GPU in use, understanding their impact is essential for accurately estimating energy consumption and evaluating the efficiency of a given configuration.

\subsubsection{Different GPUs}

\begin{figure}
    \centering
    \includegraphics[width=\linewidth]{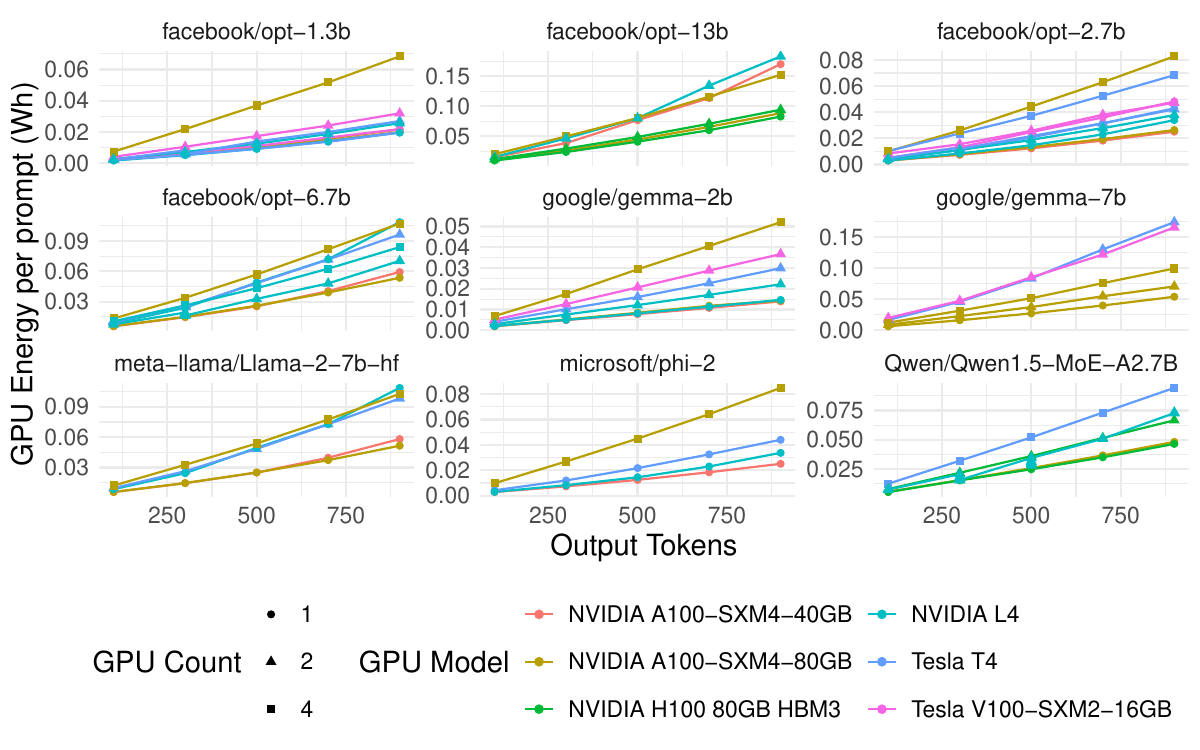}
    \caption{Energy consumption of different model with different GPUs configurations. All prompts had 300 input tokens, and 50 inputs were batched together, the results shows the average energy cost per prompt.}
    \label{fig:different_gpus_models}
\end{figure}

Figure \ref{fig:different_gpus_models} presents an experiment involving at least five distinct GPU configurations, illustrating GPU energy consumption per prompt across various language models and output token lengths. Each subplot corresponds to a different model, including Facebook's OPT series, Google's Gemma, Meta's LLaMA, Microsoft's Phi-2, and Qwen's MoE-A2.7B.

For these particular models, the configuration using four A100 80GB GPUs consistently results in the highest energy consumption. This is likely because the models are relatively small and cannot fully utilize the large memory capacity and compute resources of this setup. A more detailed analysis of when using multiple GPUs becomes beneficial is provided in Section \ref{sec:number_gpus}.

While the 4xA100-80GB setup is generally less efficient, there are exceptions. For example, in the case of the google/gemma-7b model, configurations such as 2xV100 or 2xT4 exhibit even higher energy consumption. In contrast, running these models on a single H100 or A100 GPU often delivers the best energy efficiency. For instance, the facebook/opt-13b model consumes 0.18 Wh with 2xL4 but only 0.08 Wh with a single H100.

These variations can be attributed to a combination of GPU-specific factors, including available memory, memory bandwidth, thermal design power (TDP), which ranges from 70W for the T4 to 700W for the H100, and raw computational throughput (TFLOPS).

Selecting the optimal GPU configuration is therefore critical for energy-efficient model execution. When combined with cost considerations, this selection can lead to substantial improvements in overall efficiency

\subsubsection{Varying the amount of GPUs}
\label{sec:number_gpus}

\begin{figure}
    \centering
    \includegraphics[width=\linewidth]{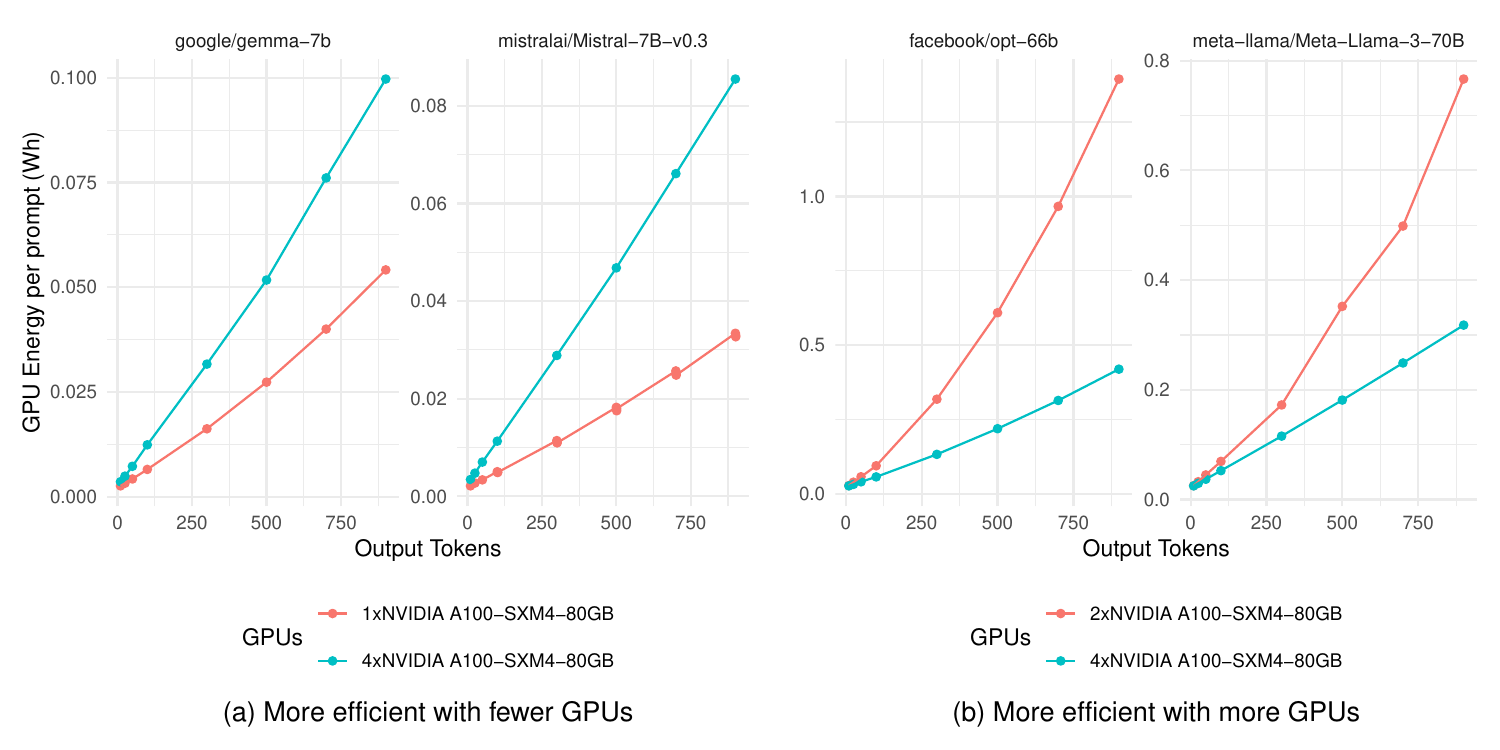}
    \caption{Energy consumption depending on the number of GPUs. Subfigure (a) shows two examples in which, with fewer GPUs, the system is more efficient, whereas (b) shows the opposite case in which having more GPUs is more energy efficient. Batch size: 50}
    \label{fig:several_gpus}
\end{figure}

Figure \ref{fig:several_gpus} presents two representative cases. In subfigure \ref{fig:several_gpus} (a), fewer GPUs lead to a more energy-efficient deployment. Our results indicate that over-provisioning GPUs within the same node for a small model does not typically yield energy savings, unless the system is subject to very high traffic demands, where such configurations might be justified.

In contrast, subfigure \ref{fig:several_gpus} (b) illustrates a scenario where a larger number of GPUs is actually more energy efficient for bigger models. This behavior is mainly due to limitations such as the available KV cache memory, which can become a bottleneck when deploying large models. In such cases, distributing the workload across more GPUs alleviates memory constraints and reduces energy inefficiencies caused by cache thrashing or data movement overhead.

\subsubsection{Acceleration with CUDAGraphs}

PyTorch offers the capability to construct CUDA graphs, which can significantly benefit smaller LLMs. These graphs mitigate the performance bottleneck caused by repeated CPU-to-GPU synchronization by pre-capturing the sequence of GPU operations and executing them as a single unit. This synchronization overhead can be particularly detrimental to the performance of smaller LLMs.

We investigated the impact of using a hybrid approach (enabling CUDA graph construction) compared to disabling CUDA graphs entirely. Disabling CUDA graphs reduces the vRAM footprint associated with graph management, potentially freeing up more memory for the KV cache.

\begin{figure}[ht]
    \centering
    \includegraphics[width=\linewidth]{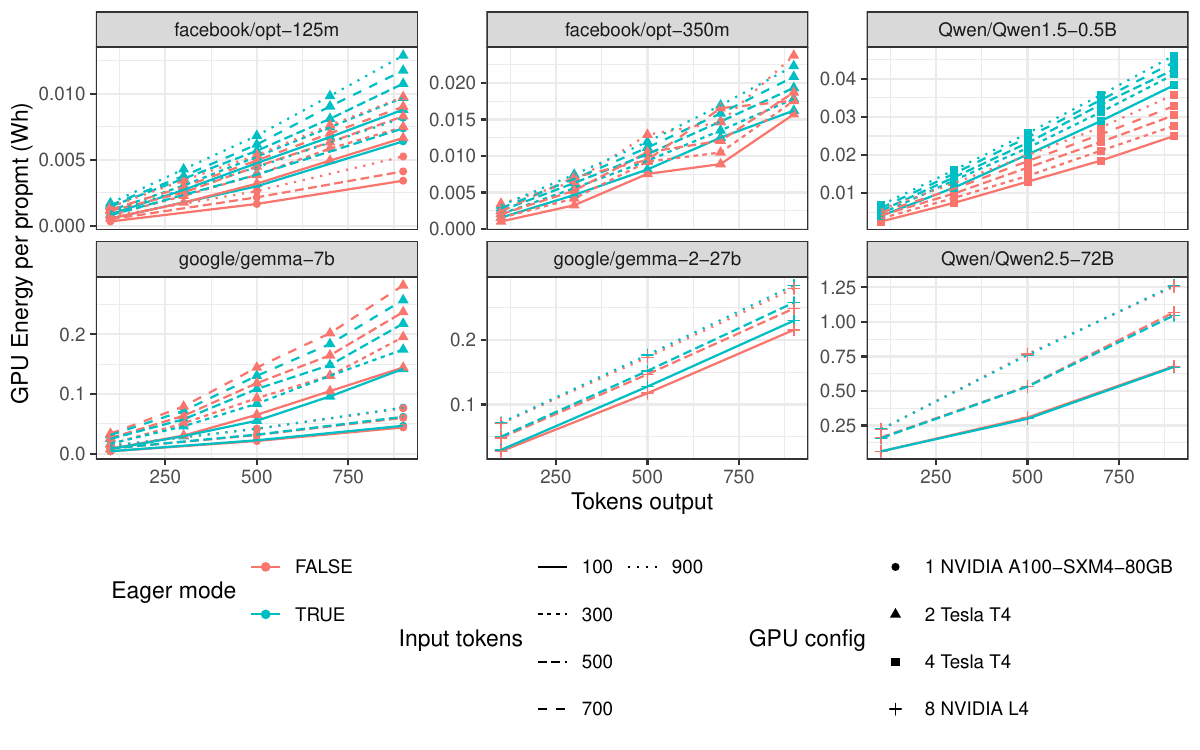}
    \caption{Energy consumption using CUDAGraphs}
    \label{fig:eager_mode}
\end{figure}

Figure \ref{fig:eager_mode} illustrates the energy consumption of various models with and without CUDA graphs enabled.

The results demonstrate that enabling CUDA graphs (i.e., deactivating eager mode) yields energy benefits for smaller models. However, with larger models like \texttt{google/gemma-7b}, the advantage of using CUDA graphs diminishes. With sufficient available vRAM, CUDA graphs can still be advantageous even for larger models, as observed with gemma-2-27b (which was run on eight NVIDIA L4 GPUs). For even larger models, the energy performance becomes less distinguishable between the two approaches.

%% file: 5.Model.tex
\section{Prediction Model}

This section presents the performance of the developed energy estimation models.

\subsection{Generalist Model}

This section focuses on the development of a generalist model designed to estimate the energy cost of a single inference. This estimation considers factors such as prompt and response length, model architecture, hardware capabilities, and current batch size.

A key requirement for this generalist model is its ability to handle batches with varying numbers and sizes of prompts. Therefore, a dedicated validation section will assess its performance under these diverse scenarios.

\subsubsection{Model Evaluation and Setup}

To evaluate the performance of the trained models, we employ repeated cross-validation. This procedure yields cross-validated R-squared ($R^{2}$), Root Mean Squared Error (RMSE), and Mean Absolute Error (MAE) metrics for each model type.

Our evaluation methodology involves isolating measurements from a specific model during the training loop. This held-out data serves as a test set to assess the model's generalization capabilities on unseen data.

The variables evaluated and used in the models are detailed in Annex \ref{annex:model_param} and are categorized into three groups: (i) model architecture; (ii) deployment hardware; and (iii) model usage (e.g., prompt length).

The models are designed to capture interactions between these variable groups. For this generalist model, we do not explicitly consider quantization due to the multitude of available methods and the lack of sufficient data to ensure accurate and specific predictions. Regarding inference quantity or batch size, we restrict this model to batches containing more than five prompts to mitigate potential overfitting.

\subsubsection{Cross-Validation Results}

Table \ref{tab:model-results} presents a comparison of the performance of six different regression models: Linear Model, XGBoost with a linear base learner (XGB Linear), standard XGBoost (XGB), Random Forest, SVM Linear, and Glmnet. The evaluation metrics used are Root Mean Squared Error (RMSE), R-squared (R2), Mean Absolute Error (MAE) and Mean Absolute Percentage Error (MAPE), as well as their standard deviations (RMSESD, R2SD, MAESD and MAPESD respectively). These metrics provide insights into the accuracy and stability of the models' predictions. We train the models with Repeated Cross Validation, with 5 folds and 5 repetitions, to find the best hyper parameters for each model. 

As shown in the table, both the XGB Linear and Random Forest models demonstrate the best overall performance. XGB achieves the lowest RMSE (0.0581) and MAE (0.0090), indicating higher accuracy. Whereas Random Forest boasts the highest R-squared value (0.9746), suggesting that it explains a larger proportion of the variance in the data. Furthermore, it provides the lowest MAPE (8.033\%). Although the Random Forest Approach also provides small errors in each of the four categories, including the most stability if terms of the Stander Deviation in MAPE (0.0980).

In summary, the results clearly favor the XGB Linear model and Random Forest, which demonstrates superior accuracy and stability compared to all other models for this task.

% latex table generated in R 4.3.0 by xtable 1.8-4 package
% Tue Jan 14 17:58:23 2025
\begin{table*}[ht]
\caption{Model Repeated Cross Validation Performance Summary. This reports the best models performance error and their standard deviation.  }
\label{tab:model-results}
\centering
\begin{tabular}{lrrrrrrrr}
  \hline
model & RMSE & RMSESD & R2 & R2SD & MAE & MAESD & MAPE & MAPESD \\ 
  \hline
Linear Model & 0.1342 & 0.0181 & 0.8326 & 0.0886 & 0.0536 & 0.0019 & 198.8234 & 9.3276 \\ 
  XGB Linear & \textbf{0.0581} & 0.0185 & 0.9710 & \textbf{0.0189} & \textbf{0.0090} & 0.0013 & 9.4908 & \textbf{0.3558} \\ 
  XGB & 0.0737 & 0.0359 & 0.9661 & 0.0214 & 0.0206 & 0.0018 & 93.0194 & 1.7550 \\ 
  Random Forest & 0.0634 &\textbf{ 0.0173} & \textbf{0.9746} & 0.0214 & 0.0091 & \textbf{0.0010} & \textbf{8.033} & 0.3814 \\ 
  SVM Linear & 0.1553 & 0.0218 & 0.8375 & 0.0544 & 0.0432 & 0.0012 & 146.2591 & 3.3791 \\ 
  Glmnet & 0.1325 & {0.0179} & 0.8524 & 0.0601 & 0.0538 & {0.0011} & 201.8731 & 4.8401 \\ 
   \hline
\end{tabular}
\end{table*}

\subsubsection{Validation with held-out LLM measurement}

To evaluate the generalization capabilities of our models, we held out a medium-sized LLM, Qwen/Qwen2.5-14B, which was excluded from the training dataset. This allowed us to assess how well the models predict energy consumption for a previously unseen LLM. The ground truth energy measurements for Qwen/Qwen2.5-14B have a median of 0.056 Wh, with a 25th percentile of 0.031 Wh and a 75th percentile of 0.094 Wh. The maximum observed energy consumption was 0.514 Wh.

Table \ref{tab:validation_qwen14} presents the validation results for the generalist models when predicting the energy consumption of Qwen/Qwen2.5-14B.

% latex table generated in R 4.3.0 by xtable 1.8-4 package
% Wed Feb  5 18:03:23 2025
\begin{table}[ht]
\caption{Validation rewults with the Generalist Model estimating the energy consumption of the model Qwen2.5 14B} 
\label{tab:validation_qwen14}
\centering
\begin{tabular}{lrrrr}
  \hline
Model & R2 & RMSE & MAE & MAPE \\ 
  \hline
Linear Model & 0.7465 & 0.0457 & 0.0378 & 158.5005 \\ 
  XGB Linear & 0.9113 & 0.0103 & 0.0070 & 17.0922 \\ 
  XGB & \textbf{0.9337} & 0.0110 & 0.0093 & 35.0943 \\ 
  Random Forest &  0.9203 &\textbf{ 0.0102} & \textbf{0.0057} & \textbf{10.4842} \\ 
  SVM Linear & 0.8542 & 0.0278 & 0.0228 & 92.1262 \\ 
  Glmnet & 0.7570 & 0.0454 & 0.0346 & 94.2163 \\ 
   \hline
\end{tabular}

\end{table}

The results clearly demonstrate the superior performance of the Random Forest model, which achieves the lowest errors RMSE (0.0102), MAE(0.0057), and more remarkable the lowest MAPE with 10.48\%. It also shows a high $R^2$ of 0.9203, indicating a strong fit to the validation data and explaining a large proportion of the variance in energy consumption.
 
In contrast, the other models exhibit considerably higher errors, with RMSE values ranging from 0.0103 to 0.0457 and MAPE values ranging from 17.09\% to 158.50\%.

While the XGB Linear model also performs strongly, achieving an $R^2$ of 0.9113 and a low RMSE of 0.0103, it does not surpass the accuracy and consistency of the Random Forest model for this validation task.

In terms of feature importance, we found that the number of GPUs, memory bandwidth, KV cache size, and the number of tokens (both input and output) are among the most influential variables in both the Random Forest and XGBoost Linear models. There are also other relevant variables such as the number of prompts (batch size), the number of parameters for the embedding or attention part of the model, as well as the number of attention heads or the number of layers, which are also relevant for understanding energy consumption.         

\subsection{Validation with Varying Prompt Sizes}
While our model performs well when all prompts within a measurement have consistent input and output token lengths, we also evaluated its performance in a more realistic setting—batches containing prompts of varying sizes. This scenario better reflects real-world usage patterns, where prompt and response lengths can vary significantly, sometimes ranging from just a few tokens to several thousand within the same batch.

To assess this, we applied the model individually to each prompt, estimating energy consumption based on its input and output token lengths. We then summed these predicted energy values and compared them to the total measured energy consumption of the entire batch. This methodology allows us to evaluate the model’s ability to generalize across heterogeneous prompt distributions.

\begin{figure}
\centering
\includegraphics[width=.8\linewidth]{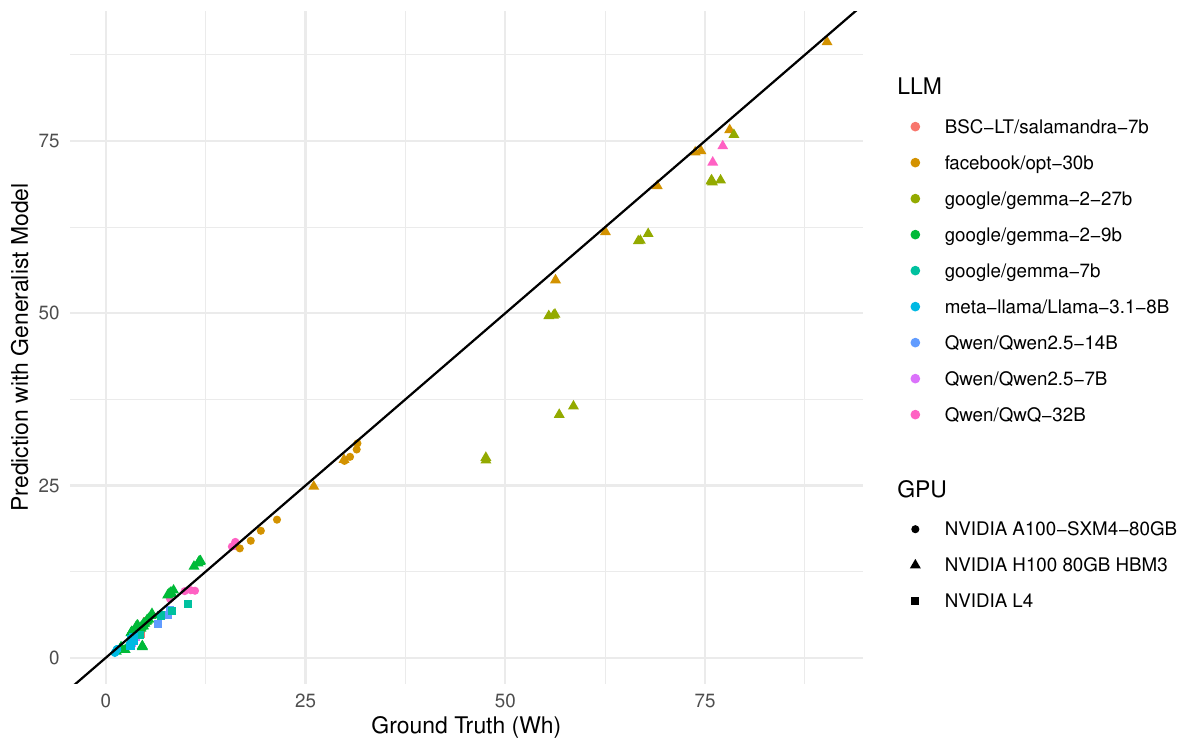}
\caption{Validation of Energy Predictions with Varying Prompt Sizes}
\label{fig:validation-plot}
\end{figure}

We conducted 125 experiments involving a variety of prompt and response lengths, batch sizes, GPU configurations, and across 9 different language models. The results, shown in Figure \ref{fig:validation-plot}, plot the measured (ground truth) total energy consumption on the x-axis and the corresponding sum of the predicted energy values on the y-axis.

The model achieved a global RMSE of 3.798 Wh when aggregating all prompts across experiments. However, normalizing this value by the number of prompts per batch reduces the RMSE to 0.0397 Wh. The corresponding $R^2$ value is 0.980, indicating a strong correlation between predicted and actual energy usage, even under significant variability in prompt sizes. Notably, the mean absolute percentage error (MAPE) remains low at 13.63\%, highlighting our approach's robustness and strong generalization capabilities.

%% file: 6.PerformanceVsEnergy.tex
\section{Performance vs. Energy}
\label{sec:performance}

While the energy consumption of Large Language Models is a critical concern, it is equally important to evaluate the quality of their responses. Choosing a model solely based on its energy efficiency is of little use if it cannot fulfill its intended task. Ideally, one should aim for a model that offers strong task performance while minimizing energy consumption.

To assess performance, various benchmarking strategies have been developed to evaluate LLMs across domains such as expert knowledge, reasoning, and general conversation quality \cite{banerjeeBenchmarkingLLMPowered2023, ivanov2024ai}. Popular benchmarks include the Massive Multitask Language Understanding (MMLU) dataset and its PRO version \cite{hendrycks2020measuring, wang2024mmlu}, WinoGrande for commonsense reasoning \cite{sakaguchi2021winogrande}, the AI2 Reasoning Challenge for scientific question answering \cite{clark2018think}, and Chatbot Arena for crowdsourced comparisons of conversational agents \cite{chiang2024chatbot}. Additionally, the LM Evaluation Harness \cite{eval-harness} supports over 60 academic benchmarks, enabling broad performance evaluations.

It is important to note that there is no single benchmark that comprehensively evaluates all capabilities of LLMs. Models may perform well in benchmarked settings but fail to generalize in real-world applications \cite{banerjee2024vulnerability}. This reinforces the need to evaluate models along multiple dimensions, including energy cost.

\begin{figure}
\centering
\includegraphics[width=1\linewidth]{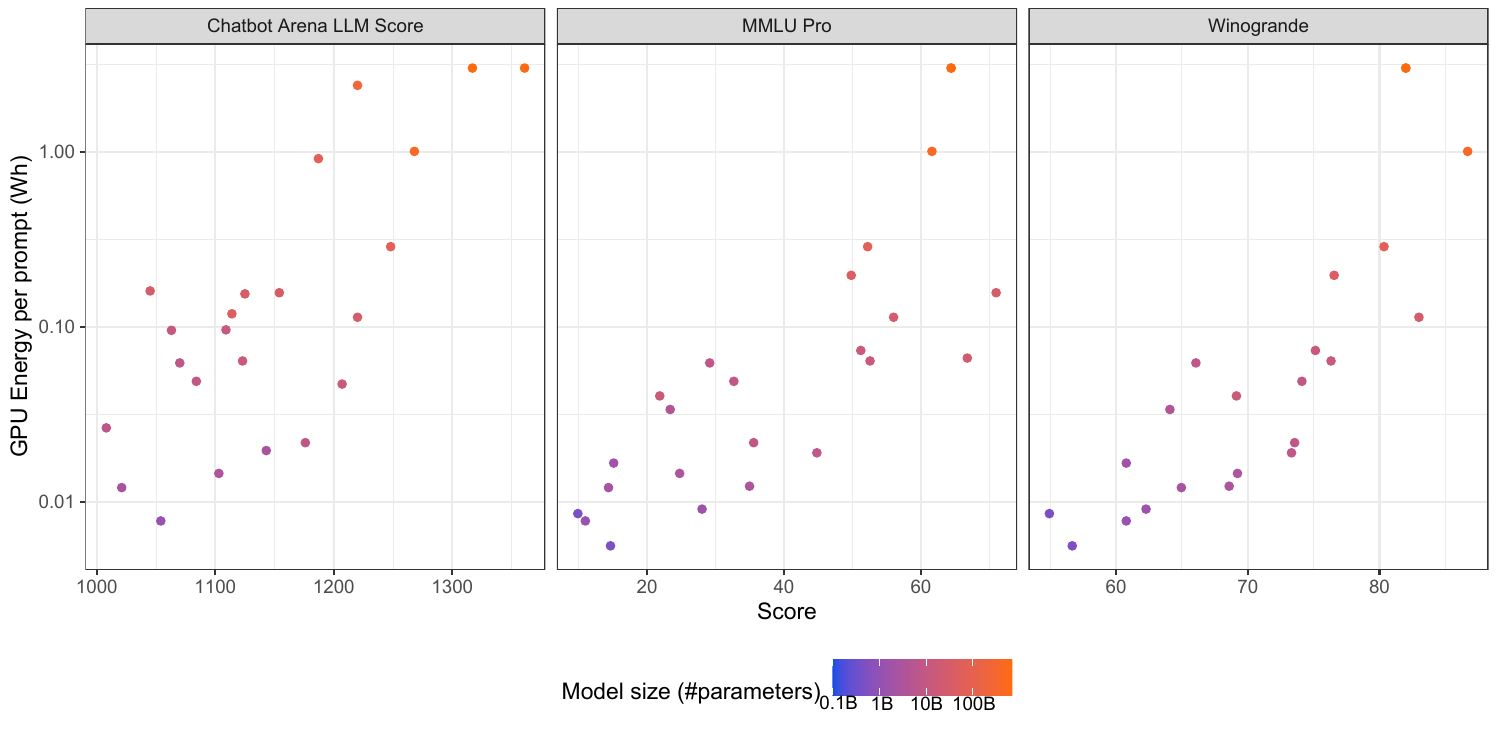}
\caption{Performance vs. energy usage per prompt}
\label{fig:perf}
\end{figure}

To explore the performance-energy trade-off, we evaluated various models on three distinct benchmarks: Chatbot Arena (human preference), MMLU PRO (expert knowledge), and WinoGrande (commonsense reasoning). Results are shown in Figure~\ref{fig:perf}, where energy usage per prompt is plotted on a logarithmic scale.

As shown in the figure, better performance typically requires greater energy expenditure. However, this trade-off is not strict. Some newer, smaller models outperform much larger ones while consuming significantly less energy. For instance, in Chatbot Arena, the Phi-4 model surpasses the Qwen2-72B model despite using roughly one-tenth the energy. This emphasizes the importance of selecting models that are not only accurate but also energy-efficient. But this trend also shows that newer, smaller models may perform better than bigger, older ones, suggesting a shift toward efficiency-oriented model design rather than unbounded scaling.

%Furthermore, public awareness of these energy costs remains limited. Most users interacting with AI services such as ChatGPT or Gemini are unaware of the underlying environmental impact. To address this, we developed a browser plugin that allows users to estimate the energy consumption of their conversations in real time. This tool aims to increase transparency and promote more sustainable usage patterns, and is described in detail in Section~\ref{sec:plugin}.

In summary, evaluating models through the lens of both performance and energy use is essential for sustainable AI. While we present initial findings here, future work should investigate these trade-offs more systematically across a wider range of tasks and deployment conditions.

%% file: 7.ConsumptionPlugin.tex
\section{Carbon-Aware Browser Plugin}
\label{sec:plugin}

To complement this research, we have developed and released a browser extension that estimates the energy consumption of conversations across various LLM platforms. The goal of the plugin is to help users understand the environmental impact, both in terms of energy use and CO2 emissions, associated with interactions on platforms such as ChatGPT, Gemini, and DeepSeek. There are other alternatives such as GPTFootprint \cite{graves2025gptfootprint}, although they have a fixed average per-query of 2.9Wh of energy, which does not acknowledge for different models neither content length. 

The plugin works by automatically detecting the exchanged text between the user and the models, identifying which LLM is being used. Using this information, we estimate the energy consumption based on several factors: the model’s approximate size (inferred from benchmarks and open-source equivalents), the number of input and output tokens, and general computational intensity. This enables us to provide a rough but informative estimate of the energy used per interaction.

\begin{figure}
    \centering
    \includegraphics[width=0.75\linewidth]{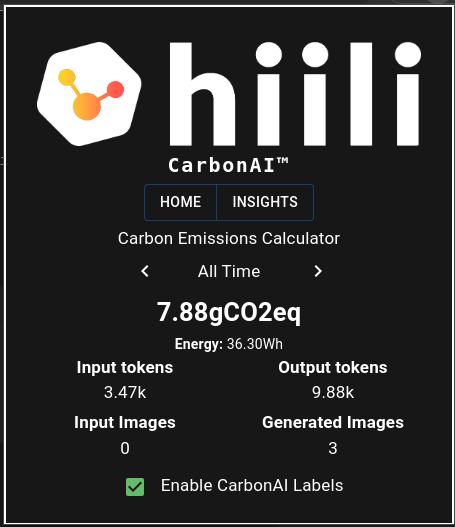}
    \caption{Screenshot of the plugin, showing how many tokens has been processed and how much energy has been consumed.}
    \label{fig:plugin1}
\end{figure}

This tool is designed to raise awareness of the environmental cost of AI usage. For example, in Figure \ref{fig:plugin1}, we present a screenshot of the plugin after it has been installed and used across multiple conversations. As shown, it displays the total number of tokens exchanged. Additionally, the plugin offers energy consumption data for each individual conversation and tags the respective platform.  

While raw energy and emission numbers may not be immediately meaningful to all users, we’ve included an Insights tab to put the data into context. This tab translates energy use into more relatable comparisons—such as the equivalent time a light bulb or television would be on, or the approximate distance one could travel by air for the same emissions. These comparisons aim to make the environmental impact more tangible and relatable.

The plugin is publicly available in the chrome store in the following URL: \url{https://chromewebstore.google.com/detail/carbon-ai-extension/clffbmdchmecickkojbefnjklbhdmmcp}

%% file: 8.Conclusions.tex
\section{Discussion}

The growing adoption of generative AI raises pressing concerns regarding sustainability. Accurately estimating the energy consumption of LLMs is a necessary step toward quantifying their broader environmental impact.

Our findings suggest that multiple factors influence the energy footprint of LLMs, even when using the same underlying model. A first consideration is usage patterns. The number of input tokens and the length of the generated output strongly affect energy requirements, as longer sequences demand more computation. Batch size also plays a crucial role: supplying a single prompt to a production system can lead to underutilized hardware resources, thereby increasing the per-prompt energy cost.

Hardware configuration is another determinant. The number and type of GPUs employed may improve or worsen energy efficiency, depending on the specific model and its memory requirements. For example, limited availability of KV cache memory can create bottlenecks, resulting in significant inefficiencies once the model is loaded. Thus, in those cases the number of prompts that can be processed simultaneously is reduced.

Model quantization offers another dimension of variability. Quantized versions may substantially reduce energy consumption in GPU memory–constrained environments. However, these benefits are not guaranteed: in some cases, performance degradation offsets potential energy savings.

Finally, architectural differences across LLMs must be taken into account. While model size correlates with energy use, it does not fully explain the disparities observed among models of similar scale. Our analysis indicates that model dimensionality influences energy consumption quadratically, whereas the number of layers contributes linearly. This underscores the need to consider not only model size but also structural design choices. Moreover, as highlighted earlier, sufficient memory availability for KV cache operations remains essential; without it, energy consumption rises sharply.

\section{Conclusions}

This work presented an extensive measurement-based study on the energy consumption of large language model inference. By systematically evaluating 155 models across 21 GPU configurations and nearly one million prompts, we provided a fine-grained understanding of how architectural features, hardware choices, and workload parameters shape energy usage. Our analysis showed that output tokens, batch size, and GPU provisioning are among the strongest determinants of energy efficiency.

Building on these findings, we introduced a predictive model that generalizes effectively across unseen architectures and diverse usage patterns. This model enables accurate estimation of per-prompt energy costs and offers practitioners actionable guidance for optimizing deployment strategies. We also explored the trade-offs between performance and energy consumption. To demonstrate practical impact, we released a publicly available browser extension that raises user awareness of the hidden energy implications of everyday LLM interactions.

Looking ahead, future work should move beyond GPU-only measurements to encompass full system-level energy profiling: a research direction we have already begun to explore, as well as deeper integration of performance metrics into sustainability assessments. As demand for LLM-based services continues to grow, our results highlight the urgency of designing systems that balance accuracy, efficiency, and sustainability. Treating energy consumption as a first-class concern will be essential to achieving more responsible and scalable AI deployments.

%% file: Annex.tex
\appendix[Energy Consumption per Model]
\label{annex:models_selected}
% \section{Energy Consumption per Model}

Table \ref{tab:model_tests} shows a table, with some of the models we have tested, showing our optimal deployment (which which GPU, number of GPUs), as well as how many parameters and energy consumption is required in an optimized way with batching (Energy), and with only one prompt (Energy$_{1}$)

% latex table generated in R 4.3.0 by xtable 1.8-4 package
% Tue Jan 14 13:09:11 2025
\begin{table*}[ht]
\caption{Summary of models tested}
\label{tab:model_tests}
\centering
\begin{tabular}{llrrrr}
  \hline
Model & GPU & GPU Count & \# Params & Energy & Energy$_{1}$ \\
\hline
deepseek-ai/DeepSeek-V3 \cite{deepseek-aideepseek-v3_2025} & NVIDIA A100-SXM4-80GB & 32 & 678 & 0.6912 & 56.9348 \\
meta-llama/Llama-3.1-405B \cite{grattafiori2024llama} & NVIDIA H100 80GB HBM3 & 16 & 405.85 & 0.3542 & 21.6722 \\
deepseek-ai/DeepSeek-V2 \cite{deepseekv2} & NVIDIA A100-SXM4-80GB & 8 & 235.74 & 0.5320 & 11.5938 \\
Qwen/Qwen2.5-72B \cite{qwen2.5} & NVIDIA H100 80GB HBM3 & 4 & 72.71 & 0.0870 & \\
meta-llama/Meta-Llama-3-70B \cite{grattafiori2024llama} & NVIDIA H100 80GB HBM3 & 4 & 70.55 & 0.0835 & \\
facebook/opt-66b \cite{zhang2022opt} & NVIDIA H100 80GB HBM3 & 4 & 65.72 & 0.1038 & \\
mistralai/Mixtral-8x7B-v0.1 \cite{jiang2024mixtralexperts} & NVIDIA A100-SXM4-80GB & 2 & 46.70 & 0.0485 & 1.1308 \\
microsoft/Phi-3.5-MoE-instruct \cite{abdinPhi3TechnicalReport2024} & NVIDIA A100-SXM4-80GB & 2 & 41.87 & 0.0700 & \\
Qwen/QwQ-32B \cite{qwq32b} & NVIDIA H100 80GB HBM3 & 1 & 32.76 & 0.0361 & 1.0753 \\
bavest/fin-llama-33b-merged \cite{Fin-LLAMA} & NVIDIA H100 80GB HBM3 & 2 & 32.53 & 0.0576 & 1.4693 \\
Qwen/Qwen1.5-32B-Chat \cite{qwen} & NVIDIA H100 80GB HBM3 & 1 & 32.51 & 0.0356 & 1.0757 \\
facebook/opt-30b \cite{zhang2022opt} & NVIDIA H100 80GB HBM3 & 2 & 29.98 & 0.0519 & 1.3640 \\
mosaicml/mpt-30b \cite{MosaicML2023Introducing} & NVIDIA H100 80GB HBM3 & 2 & 29.96 & 0.0464 & 1.2334 \\
google/gemma-2-27b \cite{gemma_2024} & NVIDIA H100 80GB HBM3 & 1 & 27.23 & 0.0289 & 0.9186 \\
mistralai/Mistral-Small-24B-Base-2501 \cite{MistralSmall3} & NVIDIA H100 80GB HBM3 & 1 & 23.57 & 0.0231 & 0.7510 \\
Vezora/Mistral-22B-v0.2 \cite{VezoraMistral22Bv02Hugging} & NVIDIA A100-SXM4-80GB & 1 & 22.24 & 0.0313 & 0.8120 \\
EleutherAI/gpt-neox-20b \cite{https://doi.org/10.48550/arxiv.2204.06745} & NVIDIA H100 80GB HBM3 & 1 & 20.55 & 0.0358 & 0.7666 \\
FelixChao/Magician-MoE-4x7B \cite{FelixChaoMagicianMoE4x7BHugging} & NVIDIA A100-SXM4-80GB & 1 & 19.73 & 0.0313 & 0.5528 \\
deepseek-ai/DeepSeek-V2-Lite \cite{deepseekv2} & NVIDIA L4 & 2 & 15.71 & 0.0363 & 0.4613 \\
Qwen/Qwen2.5-14B \cite{qwen2.5} & NVIDIA H100 80GB HBM3 & 1 & 14.77 & 0.0187 & 0.5544 \\
microsoft/phi-4 \cite{abdin2024phi} & NVIDIA H100 80GB HBM3 & 1 & 14.66 & 0.0172 & 0.4940 \\
Qwen/Qwen1.5-MoE-A2.7B \cite{qwen_moe} & NVIDIA H100 80GB HBM3 & 1 & 14.32 & 0.0152 & 0.2542 \\
Qwen/Qwen1.5-14B-Chat & NVIDIA H100 80GB HBM3 & 1 & 14.17 & 0.0238 & 0.4929 \\
microsoft/Phi-3-medium-4k-instruct \cite{abdinPhi3TechnicalReport2024} & NVIDIA H100 80GB HBM3 & 1 & 13.96 & 0.0148 & 0.4851 \\
meta-llama/Llama-2-13b-chat-hf \cite{touvron2023llama} & NVIDIA A100-SXM4-80GB & 1 & 13.02 & 0.0253 & 0.4966 \\
PygmalionAI/mythalion-13b \cite{alpinPygmalion22023} & NVIDIA H100 80GB HBM3 & 1 & 13.02 & 0.0240 & 0.4442 \\
cloudyu/Mixtral\_7Bx2\_MoE\_13B \cite{CloudyuMixtral_7Bx2_MoE_13BHugging} & NVIDIA A100-SXM4-80GB & 1 & 12.88 & 0.0235 & 0.5768 \\
facebook/opt-13b \cite{zhang2022opt} & NVIDIA H100 80GB HBM3 & 1 & 12.85 & 0.0235 & 0.5143 \\
google/gemma-2-9b \cite{gemma_2024} & NVIDIA H100 80GB HBM3 & 1 & 9.24 & 0.0147 & \\
google/gemma-7b \cite{team2024gemma} & NVIDIA H100 80GB HBM3 & 1 & 8.54 & 0.0133 & \\
meta-llama/Llama-3.1-8B \cite{grattafiori2024llama} & NVIDIA H100 80GB HBM3 & 1 & 8.03 & 0.0075 & 0.2788 \\
BSC-LT/salamandra-7b \cite{gonzalezagirre2025salamandratechnicalreport} & NVIDIA H100 80GB HBM3 & 1 & 7.77 & 0.0084 & 0.2410 \\
Qwen/Qwen1.5-7B \cite{qwen} & NVIDIA A100-SXM4-40GB & 1 & 7.72 & 0.0133 & 0.2923 \\
Qwen/Qwen2.5-7B \cite{qwen2.5} & NVIDIA H100 80GB HBM3 & 1 & 7.62 & 0.0078 & \\
mistralai/Mistral-7B-v0.3 \cite{jiangMistral7B2023} & NVIDIA H100 80GB HBM3 & 1 & 7.25 & 0.0074 & 0.2575 \\
bertin-project/Gromenauer-7B \cite{BertinprojectGromenauer7BHugging2024} & NVIDIA H100 80GB HBM3 & 1 & 7.24 & 0.0091 & 0.2645 \\
NousResearch/Yarn-Mistral-7b-128k \cite{NousResearchYarnMistral7b128kHugging2023} & NVIDIA H100 80GB HBM3 & 1 & 7.24 & 0.0092 & 0.2528 \\
meta-llama/Llama-2-7b-hf \cite{touvron2023llama} & NVIDIA A100-SXM4-40GB & 1 & 6.74 & 0.0145 & 0.2595 \\
facebook/opt-6.7b \cite{zhang2022opt} & NVIDIA A100-SXM4-40GB & 1 & 6.66 & 0.0147 & 0.2792 \\
Qwen/Qwen1.5-4B \cite{qwen} & NVIDIA A100-SXM4-40GB & 1 & 3.95 & 0.0095 & 0.1749 \\
microsoft/Phi-3-mini-128k-instruct \cite{abdinPhi3TechnicalReport2024} & NVIDIA L4 & 2 & 3.82 & 0.0137 & 0.3764 \\
meta-llama/Llama-3.2-3B \cite{grattafiori2024llama} & NVIDIA H100 80GB HBM3 & 1 & 3.21 & 0.0047 & 0.1669 \\
Qwen/Qwen2.5-3B \cite{qwen2.5} & NVIDIA A100-SXM4-40GB & 1 & 3.09 & 0.0050 & 0.1567 \\
microsoft/phi-2 \cite{MicrosoftPhi2Hugging} & NVIDIA A100-SXM4-40GB & 1 & 2.78 & 0.0074 & 0.1744 \\
facebook/opt-2.7b \cite{zhang2022opt} & NVIDIA A100-SXM4-40GB & 1 & 2.65 & 0.0073 & 0.1767 \\
google/gemma-2-2b \cite{team2024gemma} & NVIDIA H100 80GB HBM3 & 1 & 2.61 & 0.0071 & 0.2101 \\
google/gemma-2b \cite{team2024gemma} & NVIDIA A100-SXM4-40GB & 1 & 2.51 & 0.0048 & 0.1347 \\
Qwen/Qwen1.5-1.8B \cite{qwen} & NVIDIA L4 & 1 & 1.84 & 0.0049 & 0.1066 \\
Qwen/Qwen2.5-1.5B \cite{qwen2.5} & NVIDIA L4 & 1 & 1.54 & 0.0035 & 0.1281 \\
facebook/opt-1.3b \cite{zhang2022opt} & NVIDIA L4 & 1 & 1.32 & 0.0052 & 0.0854 \\
meta-llama/Llama-3.2-1B \cite{grattafiori2024llama} & NVIDIA L4 & 1 & 1.24 & 0.0028 & 0.0698 \\
Qwen/Qwen2.5-0.5B \cite{qwen2.5} & NVIDIA L4 & 1 & 0.49 & 0.0021 & 0.0677 \\
Qwen/Qwen1.5-0.5B \cite{qwen} & NVIDIA A100-SXM4-40GB & 1 & 0.46 & 0.0027 & 0.0758 \\
facebook/opt-350m \cite{zhang2022opt} & NVIDIA L4 & 1 & 0.33 & 0.0026 & 0.0533 \\
facebook/opt-125m \cite{zhang2022opt} & NVIDIA L4 & 1 & 0.12 & 0.0015 & 0.0301 \\
\hline
   \hline
\end{tabular}
\end{table*}

% \bibliographystyle{unsrt}  
% \bibliography{references}  %%% Remove comment to use the external .bib file (using bibtex).


% George Kour and Raid Saabne.
% \newblock Real-time segmentation of on-line handwritten arabic script.
% \newblock In {\em Frontiers in Handwriting Recognition (ICFHR), 2014 14th
%   International Conference on}, pages 417--422. IEEE, 2014.

% George Kour and Raid Saabne.
% \newblock Fast classification of handwritten on-line arabic characters.
% \newblock In {\em Soft Computing and Pattern Recognition (SoCPaR), 2014 6th
%   International Conference of}, pages 312--318. IEEE, 2014.

% \bibitem{hadash2018estimate}
% Guy Hadash, Einat Kermany, Boaz Carmeli, Ofer Lavi, George Kour, and Alon
% \newblock Estimate and replace: A novel approach to integrating deep neural
%   networks with existing applications.
% \newblock {\em arXiv preprint arXiv:1804.09028}, 2018.

% \end{thebibliography}

\section{Model Parameters}
\label{annex:model_param}

\begin{table*}
\large	
    \centering
\caption{Variables used to evaluate energy consumption}
\label{tab:variables_used}
    \begin{tabular}{|c|c|c|} \hline 
         \textbf{Inference Related}&  \textbf{Model Related}& \textbf{Infrastructure Related}\\ \hline 
         Input Tokens&  Number of Layers& GPU Memory\\ \hline 
         Output Tokens&  Hidden Size& GPU Bandwidth\\ \hline 
         Batch Size&  Intermidiate Size& GPU Count\\ \hline 
         Hybrid or eager mode&  Key/Value Heads& GPU TDP\\ \hline 
         KV Cache Size  per prompt&  Attention Heads& TFLOPs for current Precision\\ \hline 
         Precision&  MoE& Cuda Cores\\ \hline 
         &  Local Experts& Free vRAM\\ \hline 
 & Experts per Token&\\ \hline 
 & Layer Parameters&\\ \hline 
 & Embedding Parameters&\\ \hline 
 & Other Linear Parameters&\\ \hline 
 & Attention Parameters&\\ \hline 
 & FFN/MLP Parameters&\\ \hline 
 & vRAM requiered&\\ \hline 
 & Precision &\\ \hline
 & Quantization&\\\hline
 & Activation Function&\\\hline
    \end{tabular}

\end{table*}

Table \ref{tab:variables_used} shows a summary of the variables used to study energy estimation inferences in LLMs.